%% file: bare_adv.tex
\documentclass[10pt,journal,compsoc]{IEEEtran}

\ifCLASSOPTIONcompsoc
  \usepackage[nocompress]{cite}
\else
  \usepackage{cite}
\fi
\ifCLASSINFOpdf
\else
\fi

\usepackage{mdwmath}
\usepackage{mdwtab}
\usepackage{url}
\usepackage{epsfig}
\usepackage{helvet}
\usepackage{courier}
\usepackage{floatrow}
\usepackage{multirow}
\newfloatcommand{capbtabbox}{table}[][\FBwidth]
\usepackage{comment}
\usepackage[utf8]{inputenc}
\usepackage{amsfonts}
\usepackage{units}
\usepackage{microtype}
\usepackage{soul}
\usepackage{url}
\usepackage[small]{caption}
\usepackage{graphicx}
\usepackage{booktabs}
\usepackage{epstopdf}
\usepackage{amsthm}
\usepackage{amssymb}
\usepackage{subfigure}
\usepackage{wrapfig}

\usepackage{bbding}
\usepackage{enumitem}
\usepackage{float}
\usepackage{multicol}
\usepackage{xcolor}
\usepackage{makecell}

\usepackage{amsmath,amsfonts}
\usepackage{algorithmic}
\usepackage{algorithm}
\usepackage{array}
\usepackage[caption=false,font=normalsize,labelfont=sf,textfont=sf]{subfig}
\usepackage{textcomp}
\usepackage{stfloats}
\usepackage{url}
\usepackage{verbatim}
\usepackage{graphicx}
\usepackage{cite}

\usepackage{amssymb}

\newcommand{\new}[1]{\textcolor{black}{#1}}
\newcommand{\modified}[1]{\textcolor{black}{#1}}

\newenvironment{newtable}{%
  \begingroup
  \color{black}%
}{%
  \endgroup
}

\begin{document}
%
\title{BWTA: Accurate and Efficient Binarized Transformer by Algorithm-Hardware Co-design}
%
%
%
%

\author{Yifu~Ding, Xianglong~Liu*, Shenghao~Jin, Jinyang~Guo, Jiwen Lu
\IEEEcompsocitemizethanks{
\IEEEcompsocthanksitem Y. Ding, X. Liu (corresponding author, Email: xlliu@buaa.edu.cn) and S. Jin are with the School of Computer Science and Engineering, Beihang University, Beijing, China. 
\IEEEcompsocthanksitem J. Guo is with the Institute of Artificial
Intelligence, Beihang University, Beijing, China. 
\IEEEcompsocthanksitem J. Lu is with the Department of Automation, Tsinghua University, Beijing, China. \protect \\ }
\thanks{Our code is available at https://github.com/yifu-ding/BGEMM-CUDA}
}%

%
%

\markboth{Preprint submitted to arXiv}%
{Preprint submitted to arXiv}

%




\input{sec/0_abstract_short}

\maketitle

\IEEEdisplaynontitleabstractindextext

%
\IEEEpeerreviewmaketitle


\input{sec/1_introduction}
\input{sec/2_related_work}

\input{sec/3_motivation}

\input{sec/3_method1}
\input{sec/3_method2}
\input{sec/4_experiments}

\input{sec/5_conclusion}

\input{sec/6_acknowledgement}
\ifCLASSOPTIONcaptionsoff
  \newpage
\fi

\newpage



%

{
\bibliography{egbib}
\bibliographystyle{IEEEtran}
}




%


\begin{IEEEbiography}[{\includegraphics[width=1in,height=1.25in,clip,keepaspectratio]{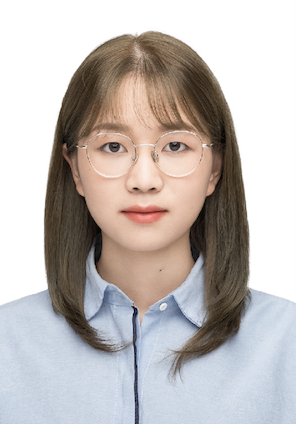}}]{Yifu Ding} is a Ph.D. Candidate under the supervision of Prof. Xianglong Liu in the School of Computer Science and Engineering \& Shenyuan Honors College at Beihang University, China. She received a B.Eng degree in Computer Science and Technology from Beihang University. Her research interest is model compression and acceleration, which is significant for AI's future with the development of big/large models.  
\end{IEEEbiography}

\begin{IEEEbiography}[{\includegraphics[width=1in,height=1.25in,clip,keepaspectratio]{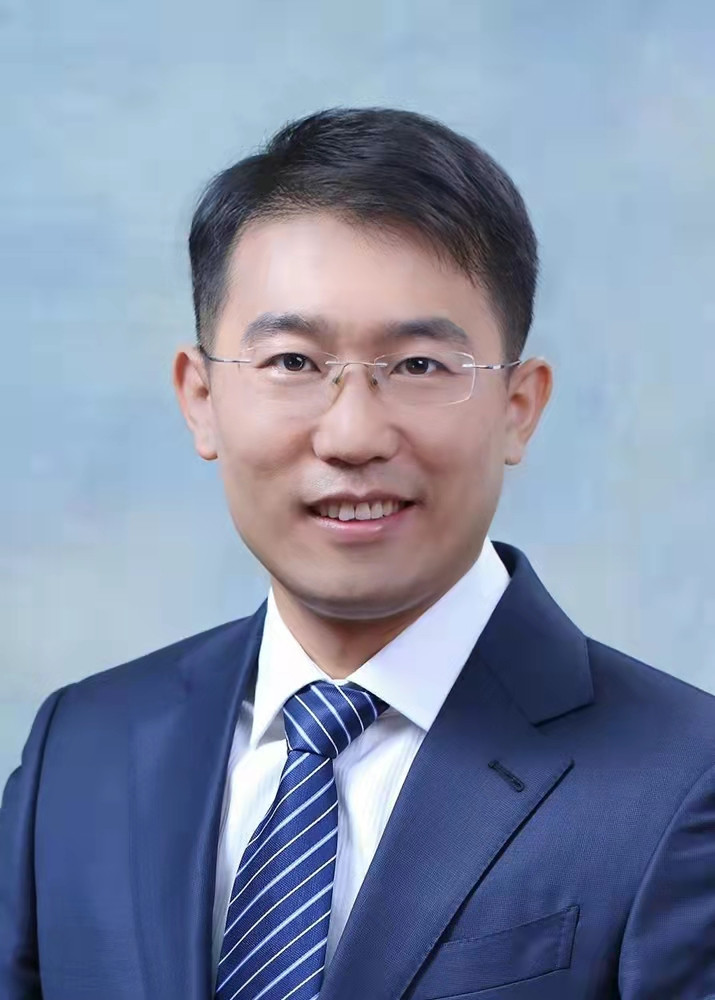}}]{Xianglong Liu}
is a Full Professor in the School of Computer Science and Engineering at Beihang University, He received BS and Ph.D. degrees under the supervision of Prof. Wei Li and visited DVMM Lab, Columbia University as a joint Ph.D student supervised by Prof. Shih-Fu Chang. His research interests include fast visual computing (e.g., large-scale search/understanding) and robust deep learning (e.g., network quantization, adversarial attack/defense, few shot learning). 
\end{IEEEbiography}


\begin{IEEEbiography}
[{\includegraphics[width=1in,height=1.25in,clip,keepaspectratio]{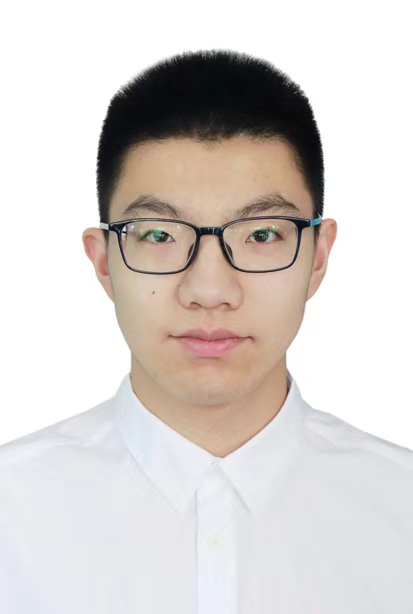}}]{Shenghao Jin}
received a bachelor's degree in computer science and technology from Beihang University and is currently pursuing a master's degree in the School of Computer Science at Beihang University. His current research interests include model quantization.
\end{IEEEbiography}

\begin{IEEEbiography}[{\includegraphics[width=1in,height=1.25in,clip,keepaspectratio]{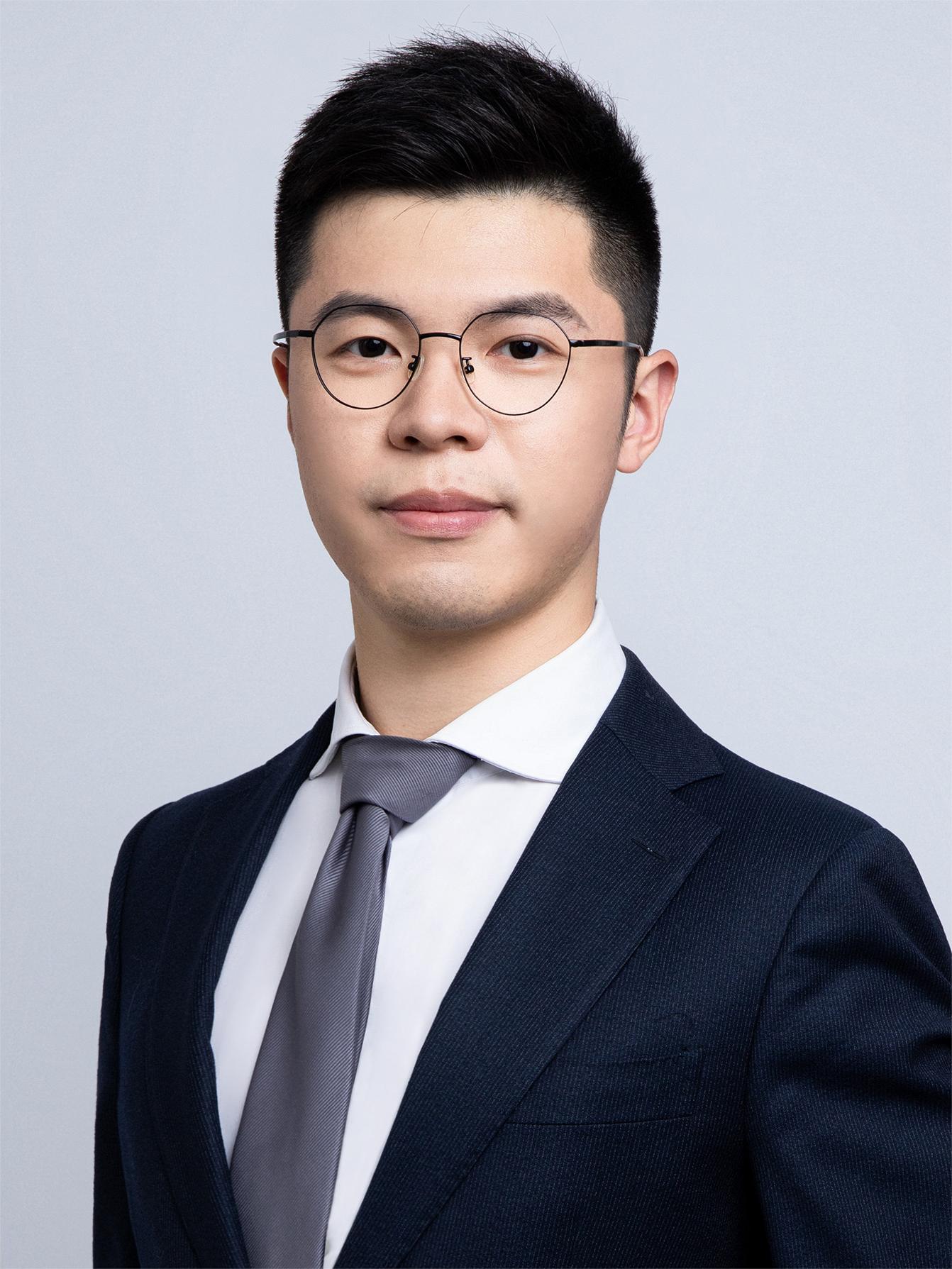}}]{Jinyang Guo}
Jinyang Guo received his Ph.D. from The University of Sydney, Australia, and got the B.E. degree (Hons.) from the University of New South Wales, Sydney. He is now an Assistant Professor at the Institute of Artificial Intelligence, Beihang University, China. His work, focuses on efficient machine/deep learning methods, has been published in top-tier journals and conferences like IEEE T-IP, CVPR, and AAAI. He also serves as a Reviewer/Program Committee Member of IEEE T-PAMI, IJCV, and so on. 
\end{IEEEbiography}

\begin{IEEEbiography}[{\includegraphics[width=1in,height=1.25in,clip,keepaspectratio]{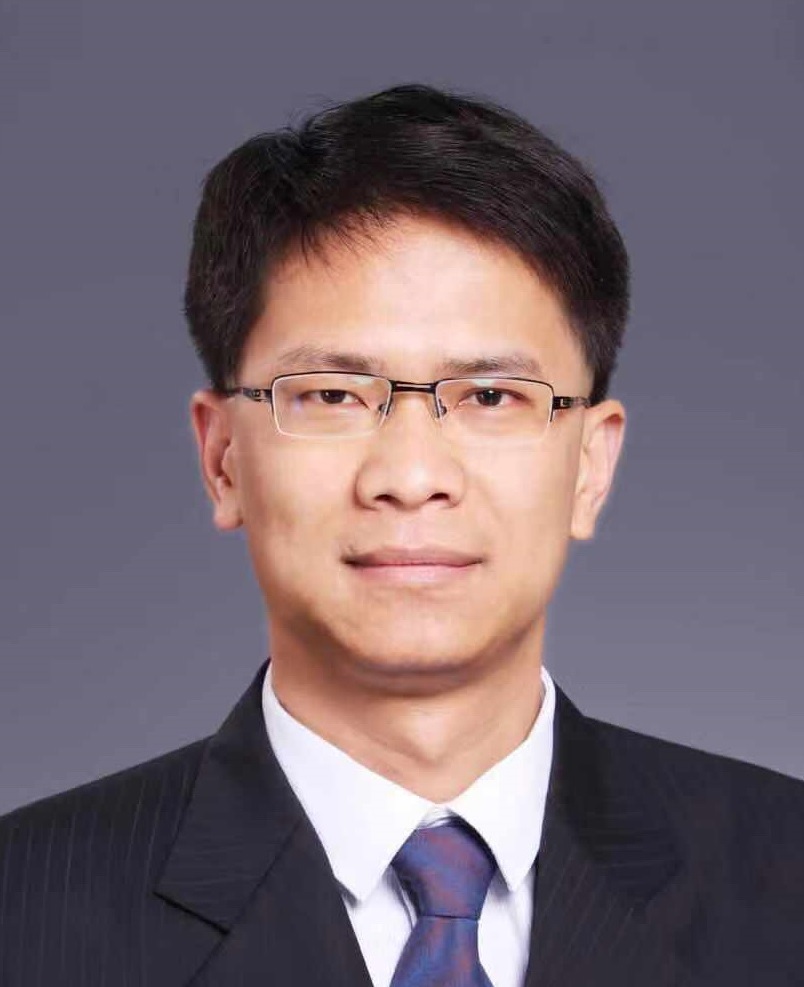}}]{Jiwen Lu}  is currently a tenured full professor and the deputy chair of the department at the Department of Automation, Tsinghua University, China. His current research interests include computer vision, pattern recognition, multimedia computing, and intelligent robotics. He was/is a member of the Multimedia Systems and Applications Technical Committee (MSA TC) and the Visual Signal Processing and Communications Technical Committee (VSPC TC) of the IEEE Circuits and Systems Society, a member of the Image, Video and Multidimensional Signal Processing Technical Committee (IMVSP TC), and so on. 
\end{IEEEbiography}




\newpage
\input{sec/appendix/7_appendix}

\end{document}

%% file: sec/0_abstract_short.tex
\IEEEtitleabstractindextext{%
\begin{abstract}

Ultra low-bit quantization brings substantial efficiency for Transformer-based models, but the accuracy degradation and limited GPU support hinder its widely usage. In this paper, we analyze the zero-point distortion in binarization and propose Binary Weights \& Ternary Activations (BWTA) quantization scheme, which projects the tiny values to zero and maintains the accuracy of extremely low-bit models. 
For training, we propose Smooth Multi-Stage Quantization, combining a Levelwise Degradation Strategy and a Magnitude-Alignment Projection Factor to enable stable, fast convergence. For inference, we develop a BWTA MatMul CUDA kernel with Instruction-Level Parallel Bitpack and comprehensive binary/ternary MatMul implementations for linear and attention operators, allowing seamless intergration across Transformer architectures. 
Experiments show BWTA approaches full-precision performance for BERT (average -3.5\% on GLUE with $<$2\% drop on five tasks), and comparable perplexity and accuracy performance for LLMs. For efficiency, it achieves 16-24$\times$ kernel-level speedups over FP16 on NVIDIA GPUs, and 216-330 tokens/s end-to-end speedup in prefill with less memory footprint on LLMs. 
As an algorithm-hardware co-design, BWTA demonstrates practical, low-latency ultra-low-bit inference without sacrificing model quality. 

\end{abstract}

}

%% file: sec/1_introduction.tex
\ifCLASSOPTIONcompsoc
\IEEEraisesectionheading{\section{Introduction}\label{sec:introduction}}
\else
\section{Introduction}
\label{sec:introduction}
\fi

\IEEEPARstart{R}{ecently}, Transformer-based models, such as the classical BERT and large language models (LLMs), have shown great potential in diverse tasks, from language, audio, vision to multimodal tasks~\cite{vig2019analyzing, van2020survey, xu2023multimodal, wang2023crossformer}. As LLMs surge in popularity, Transformers have become a research focus. 

However, the massive parameter counts and computational demands of these models lead to slow inference and substantial storage requirements, which hinder practical deployment in real-world applications~\cite{fournier2023practical, guo2021towards}. 
Researchers have proposed a variety of methods to mitigate the computation and memory burden, such as quantization~\cite{li2022qvit, guo2022squant, jacob2018quantization,frantar2022gptq}, distillation~\cite{lin2022knowledge, chen2022dearkd, wang2021knowledge}, pruning~\cite{mao2021tprune, yu2022width, cheng2024survey}, parameter sharing~\cite{panahi2021shapeshifter, reid2021subformer}, low-rank decomposition~\cite{dettmers2024qlora}, lightweight architecture~\cite{zhang2023lite, luo2022towards} and so on. 

Among these, quantization efficiently compresses the storage and accelerates the inference by compacting the parameters to lower bitwidths. 
As the most aggressive quantization scheme, binarization converts the parameters to 1-bit representation, extremely accelerates the inference by high throughput bitwise operations and compact parameter storage~\cite{qin2022bibert, 9319565, 9072484, 8444745, wang2024binaryformer, wang2023bitnet, zhu2024scalable}. 

However, there are two challenges hinder the binarization models from broad usage. \modified{(1) \textbf{Accuracy bottleneck}. Extremely low-bit parameters significantly limit the representation capability, and is rather difficult for the model to converge, which results in the accuracy drop. 
(2) \textbf{Absence of low-level support}. GPUs are the mainstream computing platform especially for LLMs. However, low-bit kernels have largely been realized only on customized hardware such as FPGAs and ASICs. These techniques are not compatible with GPUs GPUs due to architectural and resource constraints, which which hinders their deployment on GPUs and 
limits the practical value of ultra low-bit quantization. }

In this work, we begin by improving the performance of ultra-low bitwidth Transformers as the motivation, analyzing the zeropoint distortion issue in attention mechanism under binarization. \modified{Specifically, binarizing the attention scores forces a hard split, disrupting their original distribution and introducing substantial quantization error. 
Motivated by this issue, we propose the \textbf{Binarized Weight \& Ternary Activation (BWTA)}, a new bitwidth scheme for ultra-low bitwidth Transformer-based architectures that preserves the zero-point in activations. }

\input{figs/fig_overview}

We first introduce a \textbf{Smooth Multi-Stage Quantization} framework for efficient training and convergence of BWTA models. It includes a \textit{Levelwise Degradation Strategy} to gradually decrease the activation integer bins, and a \textit{Magnitude Alignment Projection Factor} that mitigates magnitude mismatch at stage transitions. This design yields smooth transitions from floating-point activations to ternary, and achieves higher accuracy and faster convergence than conventional binarization methods (see left of Fig.~\ref{fig:overview}). 

Next, to ensure the efficient end-to-end inference of BWTA Transformers, we develop a full-stack  \textbf{BWTA MatMul Kernel} on GPUs for fast execution of BWTA layers. 
\new{It fills the gap of ultra-low bitwidth inference in GPUs. 
Existing GPU binarization kernels typically binarize weights only while keeping activations in FP/INT8. For example, BitLinear~\cite{zhu2024scalable} uses floating-point activations and rewrites MatMul as additions, and recent Bitnet~\cite{wang2023bitnet} uses int8$\times$int2 linear with ternary weight and INT8 activation. Neither support low bit-width activations for further acceleration. Moreover, there is no ready-to-use, kernel-level bit-packing pipeline on GPUs that co-optimizes binary bitpacking, matrix multiplication for binary/ternary operations. (see right in Fig.~\ref{fig:overview}) }

Our kernel close this gap via efficient runtime bit-packing and native bitwise \texttt{mma}, enabling efficient binary-weights \& ternary-activations computation in GPUs. 
It has two key components: \textit{Instruction-level Parallel Bitpack}, which converts floating-point tensors to binary or ternary representations on the fly, and 
\textit{Comprehensive MatMul implementations} that support the arithmetic rules required by binary/ternary operands for both linear and attention, bringing broad compatibility with Transformer-based models.

Comparing with existing FP and low-bit linear ops, our BWTA kernel achieves significantly lower latency, which is approximately 23.6$\times$ speedup for linear, 16.7-18.5$\times$ for the attention compared to the FP16 counterpart with typical feature dimensions. It is further validated in end-to-end inference on 2B-parameter LLMs, achieving 216–330 tokens/s during prefilling and 12–15 tokens/s during decoding, outperforming the previous SOTA kernel.

\modified{We present, to the best of our knowledge, the first binary-weight \& ternary-activation (BWTA) quantization scheme for Transformer-based architectures, as well as the first end-to-end system support for ultra-low bitwidth inference on GPUs. It maintains model performance by smooth multi-stage training strategy, and can be seamlessly integrated into various transformer-based models and structures to enjoy the speedup benefits. Our work paves the way for ultra-low bit quantization for Transformer-based architectures, demonstrating its practical utility through accuracy performance and low inference latency. }


\modified{We summarize our main contributions of this paper as:
\begin{itemize}
    \item A binary-weight \& ternary-activation (BWTA) framework with Smooth Multi-Stage Quantization strategy for stable training; 
    \item A state-of-the-art customized BWTA MatMul CUDA kernel, providing 16-24$\times$ speedup vs. FP16 counterpart, and brings leading end-to-end acceleration in both prefill/decode phases; 
    \item The accuracy of the BWTA Transformer significantly outperforms that of conventional binarization methods on both BERT and LLM-based architectures. 
    \item BWTA is broadly compatible with transformer-based architectures, and can be seamlessly integrated to enjoy the speedup benefits. 
\end{itemize}}

%% file: figs/fig_overview.tex
\begin{figure*}[t]
 \vspace{-0.2em}
    \centering
    \includegraphics[width=\linewidth]{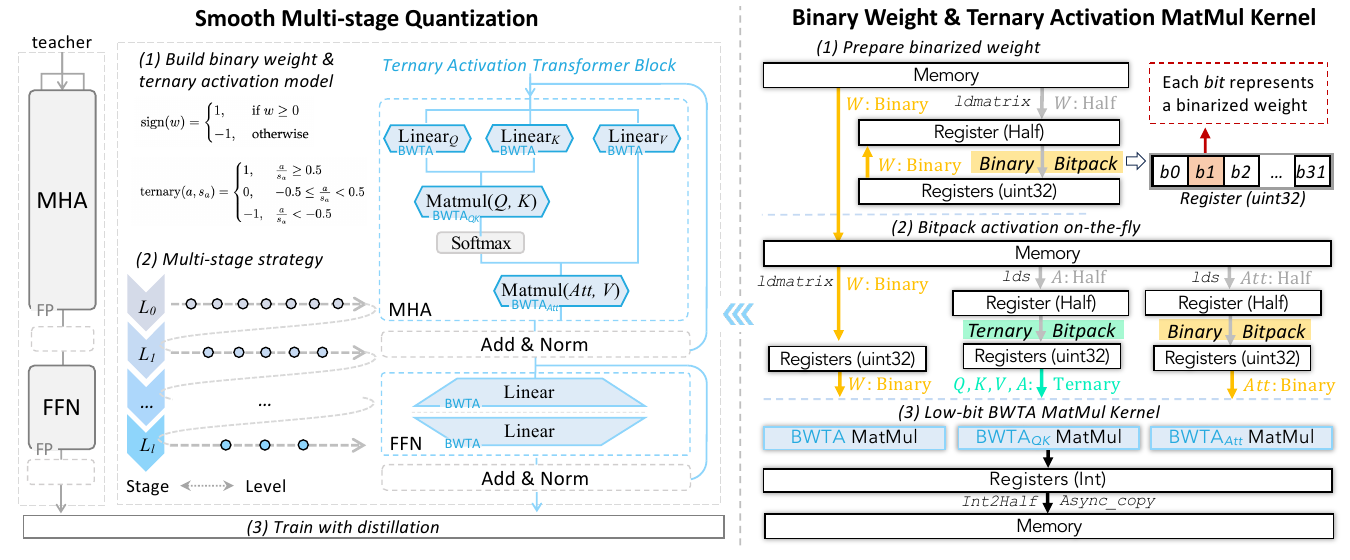}
    \caption{The overview of the Binary Weight \& Ternary Activation (BWTA) framework. The left is the training algorithm named \textbf{Smooth Multi-stage Quantization}, designed for ultra-low bit Transformers with ternary activation, including the (1) binary/ternary definitions to build a BWTA model, (2) smooth strategies for stable convergence, and (3) distillation for fast training. The right is the full stack GPU support for the custom \textbf{BWTA MatMul Kernel}, including the (1) binarized weight for storage reduction, (2) instruction-level parallel bitpack for runtime quantization, and (3) low-bit BWTA MatMul kernels for fast computation. } 
    \label{fig:overview}
     \vspace{-0.6em}
\end{figure*}

%% file: sec/2_related_work.tex
\section{Related Work}  

\textbf{Binarization for Small-scale Transformer-based Models. }
Binary quantization, which compacts the parameters to only 1-bit, can significantly compress the model size, saving the storage and memory footprint on devices
\cite{9319565, 9072484, 8444745, wang2024binaryformer, bai2021binarybert, qin2022bibert, liu2022bit, mlbert, xing2024bipft, bebert}. \new{It has also been extended to many Tranasformer-based variants and downstream tasks, bringing valuable insight and impressive performance. For example, Liu et al.~\cite{liu-2023-binary-and-ternary-nlg} extend to BART for natural language generation, Le et al.~\cite{binaryvit} and He et al.~\cite{bivit} extend binarization to ViT for vision tasks. }

\textbf{Binarization for Large Language Models. }
\new{With the rise of large language models (LLMs) and their substantial computational demands, low-bit quantization has attracted significant attention. Representative methods include GPTQ and AWQ for post-training quantization, and Q-LoRA for accelerating training. Recently, 1-bit LLM approaches have also emerged: DB-LLM~\cite{chen2024dbllm} employs dual binarization by using two 1-bit matrices to approximate each linear weight matrix; PB-LLM~\cite{shang2023pbllm} replaces selected layers with pure binarization to reduce cost while retaining stability; and BiLLM~\cite{huang2024billm} further compresses the average weight precision to nearly 1.1 bits through structured binarization and calibration. 
These weight-only methods substantially reduce model storage, highlighting parameter redundancy in LLMs and the promise of 1-bit quantization. }

\modified{However, the performance of ultra low-bit quantized models are always degraded, even when keeping the activations in LLMs in BF16. Also, ultra-low bitwidth neural network is hard to train and converge, which is well recognized in previous studies~\cite{Nagel2022Overcoming, DSQ}. 
Therefore, in this work, we propose to quantize activations to ternary based on our observations, and quantize activations to ultra low-bit for the first time in LLMs. }


\textbf{Hardware Support for Quantization. } 
On GPUs, many inference frameworks propose the state-of-the-art kernel supports for low-bit quantization, such as TensorRT-LLM\footnote{https://docs.nvidia.com/tensorrt-llm}, Transformers by HuggingFace\footnote{https://huggingface.co/docs/transformers/v4.44.2/quantization}, vLLM\footnote{https://github.com/vllm-project/vllm}, bitsandbytes\footnote{https://github.com/bitsandbytes-foundation/bitsandbytes} and so on. Actually, most frameworks only support INT4 or INT8 data types for MatMul kernels in underlying implementation. 
The newly proposed works on LLMs have emerged GEMV-based low-bit MatMul on NVIDIA GPUs~\cite{Atom, BiQGEMM}. \modified{Zhu et al.~\cite{zhu2024scalable} propose a ternary weight BitLinear operator to replace MatMul with additions and negations to reduce the compute and memory, while keeps the activations to floating-points. BitNet~\cite{wang2023bitnet} also designs custom linear ops receiving INT2 weight with INT8 activation for faster linear layer. These kernels are pioneer works in ultra-low bit implementation for LLMs, which both deal with ternary weight and higher bitwidth support for activations. Till now, there is no ready-to-use bitpacking methods and MatMul kernels available for the ultra-low bitwidth. }

Faster low-level implementations on FPGAs or ASICs have been widely studied~\cite{bilaniuk2019bit, taka2024efficient}. FINN~\cite{umuroglu2017finn} officially released by Xilinx provides a deployment of binarized neural networks, which support fully binarized (W1A1) layers for convolution and linear layer, which shows far less system power usage and extremely low latency. \new{Also, works such as SoftmAP~\cite{rakka2024} and SOLE~\cite{sole} designs customized integer approximations for nonlinear operations, which solves the issue of time-consuming floating-point computation on the chips. } We believe that ultra-low bit quantization is a promising technique for compressing large models. 

However, the techniques in FPGAs or ASICs can not be directly applied to GPUs due to the different hardware architecture and resource configurations. Therefore, our work fills the blank in ultra-low bitwidth inference on GPUs for both weight and activations. By providing foundational system support for datatype conversion (packing floating-point numbers to binary or ternary) and dense computation (matrix multiplication), our kernel is a comprehensive and easy-to-use operator for real speedup and broader implementation.

%% file: sec/3_motivation.tex
\input{figs/fig_motivation}

\section{The Rationale of Ternary Quantization}

\modified{The attention mechanism is the fundamental component in transformer-based architectures. 
One thing that should be mentioned is that, since the attention probabilities are calculated by $\operatorname{Softmax}$, the output obeys the power-low distribution, which means there are most small values around zero while seldom large values. Therefore, if we adopt the $\operatorname{bool}$ function as the common practice in ~\cite{qin2022bibert, liu2022bit}, most of the attention probabilities become zero, resulting in the attention outputs cluster around zero. If we apply $\operatorname{sign}$ function as binarization for activations in the next layer, following traditional binarization framework, there will be an inevitable \textit{zeropoint distortion} issues. }

We show the actual parameter distribution in Fig.~\ref{fig:ternary_motivation} to provide an emperical evidence on the above issue.
\modified{The Y-axis is the value of the activation variable, and the X-axis is the training epoch. Grey histograms are the activation before quantized (full-precision), Green histograms in (a) are \textbf{Bitwise} quantized parameters for value $V$ and attention score $A$ under bitwidths $k=3,2,1$, and blue histograms in (b) \textbf{Levelwise} are ones using multi-stage quantization strategy proposed in Sec.~\ref{sec:method1} using $9,7,5,3$ levels. }

\modified{See Fig.~\ref{fig:ternary_motivation}(a), $V$ before quantization is zero-mean, with lots of relatively small values around zero. It is also a typical distribution for most activations in linear. Before binarization at epoch 15, there is an integer to represent 0, which accounted for about 57\% parameters, while other integers are 22\%, 19\% and 2\%. However, when transiting to binary at 15 epoch. The activation degrades to double peak in half, eliminating zero in representation. It brings significant quantization error since the tiny values around zero are forcefully divided into two parts. This also exists in the MatMul operation, $A_{\mathrm{bitwise}}$ has 52\% values quantized to zero at the 2-bit stage, while being eliminated as well when transiting to binary in the immediate binary layer. }
\modified{Meanwhile, since there are even levels ($2^k$) in bitwise activations, if always let one level present zero, the number of integers on positive and negative axes are unequal.
Therefore, we propose to apply ternary quantization to activations, which mitigate the zeropoint distortion problem by always keeping zero to project tiny values when being quantized, making the distribution consistent with the full-precision. }

However, two major challenges come to mind: \textit{How to train a BWTA network efficiently?} \modified{Can it achieve real acceleration in inference? To put it into realistic, we build the BWTA model with $\operatorname{sign}$, $\operatorname{bool}$ and $\operatorname{ternary}$ quantization functions, with the details in Appendix.~\ref{sec:baseline_scheme}. }
In the following sections, we detail the \textbf{Smooth Multi-stage Quantization} framework for efficient training and fast convergence in Sec.~\ref{sec:method1}. Then, we elaborate the custom \textbf{BWTA MatMul Kernel} to support the fast computation in inference in Sec.~\ref{sec:method2}.

%% file: figs/fig_motivation.tex
\begin{figure}[t]
    \centering
    \includegraphics[width=0.95\linewidth]{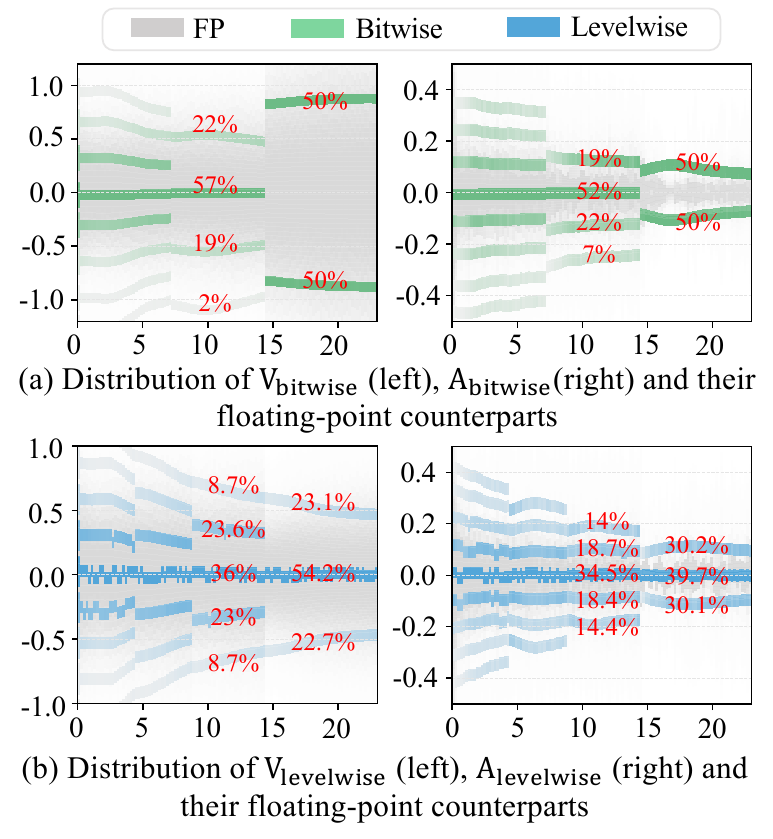}
    \caption{Histograms for binary/ternary activation in Self-Attention structure before and after quantized. (a) shows the value matrix $V_\mathrm{bitwise}$ and the activation $A_\mathrm{bitwise}$ in the linear layer following the multiplication of attention score and value trained by bitwise degradation strategy, while (b) shows the distribution of $V_{\mathrm{levelwise}}$ and  $A_\mathrm{levelwise}$ by levelwise strategy.}
    \label{fig:ternary_motivation}
     \vspace{-0.6em}
\end{figure}

%% file: sec/3_method1.tex
\section{Smooth Multi-Stage Quantization}
\label{sec:method1}

To achieve fast convergence and stable optimization BWTA models, we propose a \textbf{Smooth Multi-stage Quantization} training framework that progressively transits the parameters from the continuous real numbers to the extreme discrete ternary space. It includes two key techniques: 
(1) \textbf{Levelwise Degradation Strategy}, which employs an equal distribution of integer levels around zero to address the zeropoint mapping issue, and a progressive transition strategy to mitigate the sudden drop of representation capacity as the integer space contracts, 
(2) \textbf{Magnitude Alignment Projection Factor}, which aligns the value magnitudes between consecutive stages, preventing the drastic change of value magnitudes and facilitating a smoother transition between stages. 
As shown in the left part in Fig.~\ref{fig:overview}, we first build a BWTA model as described in Appendix.~\ref{sec:baseline_scheme}, 
and train the quantized model with knowledge distillation. 

\input{figs/fig_zeropoint}

\subsection{Levelwise Degradation Strategy}

Previous multi-stage quantization approaches that use the integer range $[-2^{(k-1)}, 2^{(k-1)}-1]$ ($k$ is the bitwidth) face two major issues when mapping the floating-point zero ($0_{fp}$) to the limited integer space. We illustrate in Fig.~\ref{fig:2practices}.

\modified{\textit{Practice 1} without zeropoint $\mathrm{zp}_{fp}$ will encounter the level unbalancing issue. Since the total number of integers after quantization is even, it creates an imbalance in the number of integer bins on positive and negative axes, leading to an uneven capacity of information and distribution mismatch compared to original floating-point model. }

\modified{\textit{Practice 2} uses $\mathrm{zp}_{fp}$ possibly encounters the level unbalancing issue ({Issue 1}) or no direct mapping of $0_{fp}$. $0_{fp}$ is shifted by $\mathrm{zp}_{fp} \not\in \mathbb{Z}$ to a real number between two integers, and then rounded to the nearest integer. Given that the activations are Gaussian-like with zero-mean, the large amounts of tiny values around zero can accumulate significant quantization error. }

\begin{figure}
    \centering
    \includegraphics[width=1.0\linewidth]{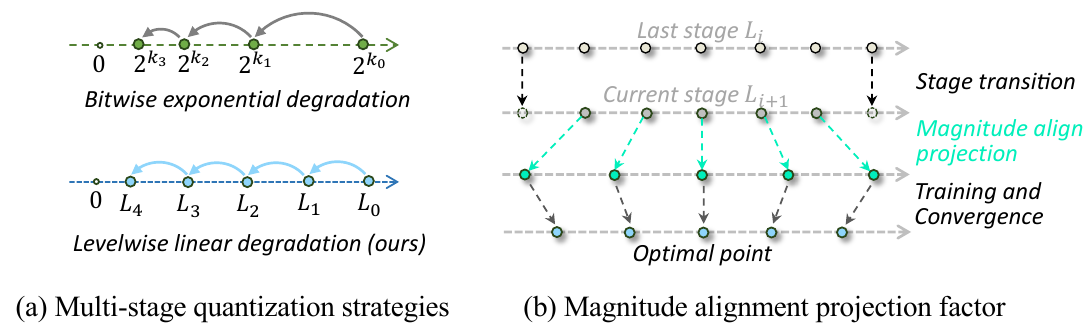}
    \caption{(a) The illustration of the bit/levelwise multi-stage quantization strategies. (b) The grid points shift with projection factors (blue arrows), and convergent again in the next stage with fewer integers (gray arrows). }
    \label{fig:method1}
\end{figure}

Therefore, we propose a novel \textbf{Levelwise Degradation Strategy} for ultra-low bitwidth training. It applies an equal number of quantized levels at both axes. We set the range $[-L, L]$, where $L\in \mathbb{Z}^+$ for quantization bins. $L$ starts from an arbitrary positive integer $L_0$ at the initial stage. 
Subsequently, the range of quantized integers decreases smoothly in a descending sequence $\{L_0, L_{1}, \dots, L_{l}\} \in\mathbb{Z}^+$, where $l \ge 0$ denotes the stages. Making sure that the $1=L_l<\dots<L_{2}<L_{0}$, so that the quantization space is continuously decreasing. This strategy is more flexible than the traditional power of two quantization in two aspects: it starts from an arbitrary integer instead of $2^k$, and it decreases by a smaller step size rather than halving at each stage. 

See Fig.~\ref{fig:method1}(a) for illustration. Traditional bitwise degradation shrinks the quantization space exponentially, e.g., $\{16, 8, 4, 2\}$. While our levelwise degradation strategy is more gradual, bridging adjacent stages with smaller gaps, e.g., $\{17, 15, 13, \dots, 3\}$. It avoids the drastic change in information granularity and capacity during stage transitions. The training epochs for each stage can be evenly distributed or use an early-stop strategy once it reaches a satisfactory loss landscape. Empirically, we allocate half of the epochs to the final (ternary) stage ($L_l$), and evenly divide the remaining epochs for the other stages. 

\subsection{Magnitude Alignment Projection Factor}
\label{sec:magnitude_alignment}
Although the levelwise decreasing is more gentle than the bitwise one, there still exists a reduction in the number of integer levels. It causes the sharp accuracy drop at stage transition, as shown in ablation experiment Sec.~\ref{sec:ablation_multistage}. 

Therefore, we superpose an efficient projection factor $p$ at the transition, applied to the learnable scaling factors to recover the magnitude when transitioning to smaller integers. The factor $p$ is simply defined by the ratio of the accumulated absolute values of the activations in two adjacent stages: 
\begin{equation}
\label{eq:projection_factor}
    s_A^{i} = s_A^{i-1} \cdot p = s_A^{i-1} \frac{\sum(|{A_{i-1}}|)}{\sum(|{A_i}|)}, \quad i \in [0, l], 
\end{equation}
where $s_A^i$ and $s_A^{i-1}$ are the scaling factors for activation at the $i$-th (current) and the $(i-1)$-th (last) stages. 

\modified{Fig.~\ref{fig:method1}(b) illustrates the projection factor. 
It enables a smoother transition between adjacent stages.
After a few training epochs, the scaling factors converge again to their optimal value. 
We conduct the experiment in Sec.~\ref{sec:ablation_factor} to show the effect of the projection factor incorporating with the scaling factors on smoothing the loss curves and stabilizing the parameter optimization. }

%% file: figs/fig_zeropoint.tex
\begin{figure}[t]
    \centering
    \includegraphics[width=0.97\linewidth]{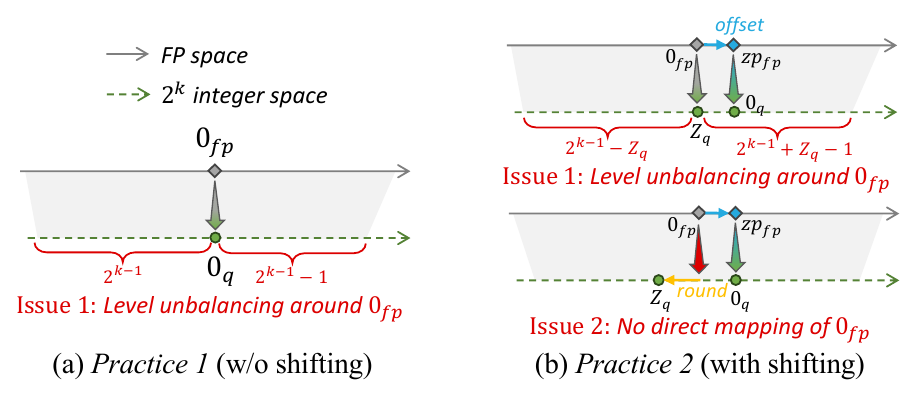}
    \caption{Illustrations for two practices and the issues of projecting $0_{fp}$ to integer space. }
    \label{fig:2practices}
     \vspace{-0.6em}
\end{figure}

%% file: sec/3_method2.tex
\section{Binarized Weight and Ternary Activation MatMul Kernel}
\label{sec:method2}
To ensure fast inference of BWTA Transformers, we develop a full-stack CUDA kernel for the efficient execution of BWTA layers, including the linear and attention ops, making it broadly compatible with most Transformer-based models and architectures. We incorporate two major techniques:
(1) \textbf{Instruction-level Parallel Bitpack} for on-the-fly quantization of variables to binary or ternary formats, enabling real-time inference, and 
(2) \textbf{Comprehensive Computational Implementations} to support different arithmetic rules, ensuring compatibility with various Transformer-based models. 
The carefully engineered MatMul kernels guarantee the efficiency and versatility of the BWTA quantization, providing an promising quantization paradigm for the ultra-low bit Transformer-based models.

\subsection{Adaptations to Wheels}
First, we continue the useful practices in FP6-LLM~\cite{xia2024fp6} with necessary modifications, which can be roughly summarized in four major aspects: 

\noindent\textbf{Double buffer at register level}: During each MMA instruction computation, the input for the next iteration is loaded into the buffer register and conducted bitpacking process to avoid the conflicts of read and write. Unlike FP6-LLM, we only apply a double buffer to weights, while activations in each thread block are bitpacked once and stored in advance. 

\noindent\textbf{Quantization and dequantization processes}: It requires runtime quantization instead of dequantization before the MMA instruction. As we use actual binarized GEMM in our implementation, the input matrices should be quantized to bitstreams with each bit representing a binary/ternary value. The results can be converted back from int32 to floating-point to keep the consistency of the dataflow before being written back to the GPU's shared memory. 

\noindent\textbf{Local processing of activation with SIMT core}: Activations are processed locally within each thread using SIMT core, which will not be written back to the shared memory but stored in registers for future use, eliminating unnecessary round-trip accesses to shared memory. 

\noindent\textbf{Data Loading with $\texttt{lds}$ instructions}: We employ \texttt{lds} to load both weights and activations from shared memory to registers instead of using \texttt{ldmatrix}. The \texttt{lds} is flexible with fine-grained data layouts, allowing for future scalability. 


\subsection{Design Methodology}

\begin{figure}
 \vspace{-0.6em}
    \centering
    \includegraphics[width=\textwidth]{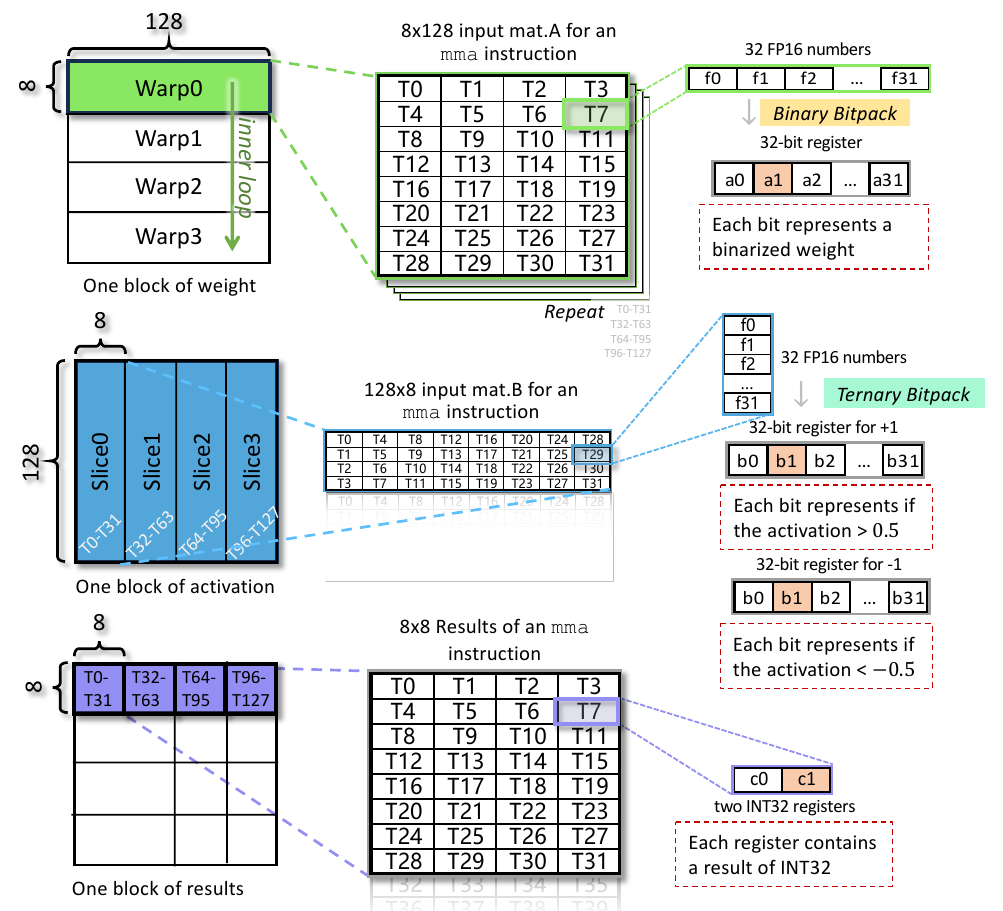}
    \caption{Data distribution in threads for weight, activation and \texttt{mma} results using \texttt{m8n8k128} data layout. Binary weight is packed and stored before inference, and we pick $8\times 128$ elements for each SIMT, with four copies stored in 128 threads. Activation is packed into ternary during runtime with four slices each SIMT without repeated copies. } 
    \label{fig:thread_allot}
     \vspace{-0.6em}
\end{figure}



\subsubsection{Data Allocation for bitwise \texttt{mma} instruction} 

Both input matrices of MatMul should be carefully allocated to registers in each threadblock in the correct layout.  

We give an example of \texttt{mma} instruction with \texttt{m8n8k128} data layout in Fig.~\ref{fig:thread_allot}, which receives input matrices of sizes $8 \times 128$ and $128 \times 8$. Assuming the weight and activation matrices are $32 \times 128$ and $128 \times 32$, if we run 128 threads, they can be chunked into four $8 \times 128$ partitions. We run an inner loop for iterating the weight partitions.
Each threadblock picks the same warp of weight and then stores them into the corresponding shared memory address. In other words, we load the weight as four copies. The inner loop will iterate the warps of weight by SIMT procedure. 
For activations, we divide the data matrix into four slices as well, with each slice processing $128 \times 8$ elements. Each group of 32 threads executes an \texttt{mma} instruction of \texttt{m8n8k128} data layout, processing $8\times 128$ weight and $128 \times 8$ activation. 

Finally, we get four $8\times8$ elements as a result. As shown in the purple squares in Fig.~\ref{fig:thread_allot}. 
The results are stored in each thread with two INT32 registers. Before being written into the shared memory according to their coordinates, they can be converted to floating-point to maintain consistency with the outside dataflow.

\subsubsection{Instruction-level Parallel Bitpack}
\label{sec:bitpacking}
\begin{figure} 
    \centering
    \includegraphics[width=\textwidth]{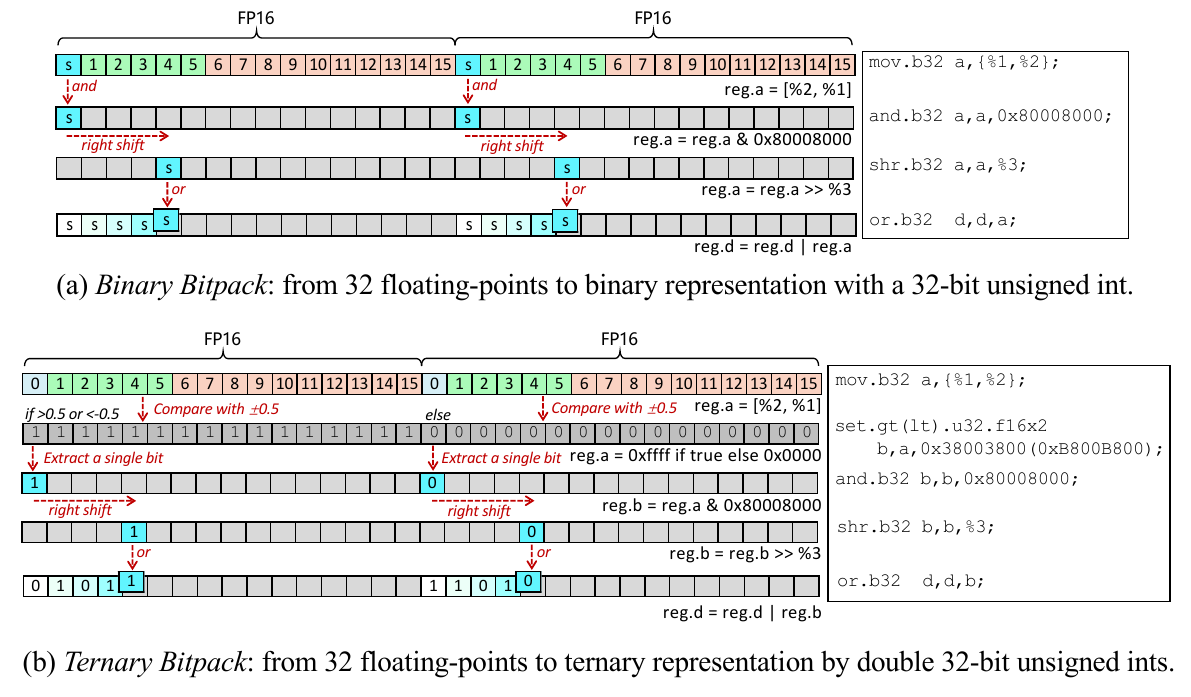}
    \caption{Instruction-level parallel bitpack from 32 FP16 (Half) to (a) binary and (b) ternary representations. }
    \label{fig:bitpacking}
    \vspace{-0.6em}
\end{figure}

Since \texttt{mma} instructions with bitwise operations (i.e. \texttt{and} or \texttt{xor}) accept unsigned bitstrings (such as unsigned INT32) as inputs, we need to convert floating-points activation into 1-bit formats on-the-fly to ensure runtime inference. 
To facilitate faster bitpacking, we design an Instruction-level Parallel Bitpack, as illustrated in Fig.~\ref{fig:bitpacking}. 

\textit{Binary Bitpack} packs every 32 floating-points to unsigned integer registers, with each bit representing one FP value. As shown in Fig.~\ref{fig:bitpacking}(a), first, we pick two FP16 values with an interval of 16, where \texttt{\%3} is the index of the two values in a group of 32 values. Then we store them into 32-bit register \texttt{reg.a} using \texttt{mov}, then use \texttt{and} for the two FP16 values and \texttt{0x80008000} to extract the signbits, replacing the original value in \texttt{reg.a}. Then, we do a right shift on \texttt{reg.a} by \texttt{\%3} bits using \texttt{shr}. The results are accumulated into a target 32-bit register \texttt{reg.d} by \texttt{or}, and then proceed to the next 32 values. Notably, negative numbers are stored as bit \texttt{1} in the register, while positive numbers are stored as bit \texttt{0}. 

\modified{\textit{Ternary Bitpack} has similar process as the binary bitpack, but it compares the FP16 values against \texttt{0x38003800} as the hex format of double $0.5$ in FP16, and \texttt{0xB800B800} as that of double $-0.5$ using \texttt{set}. If the FP16 number is greater than $0.5$ (or is below $-0.5$), the result is \texttt{1} for 16-bit; and \texttt{0} for 16-bit otherwise. We apply \texttt{and} to the result register \texttt{reg.b} with \texttt{0x80008000} to extract the results. We need two 32-bit registers to store the results in ternary bitpack. }

\subsubsection{Comprehensive Computational Implementations}


\begin{figure}
     \vspace{-0.6em}
    \centering
    \includegraphics[width=0.95\textwidth]{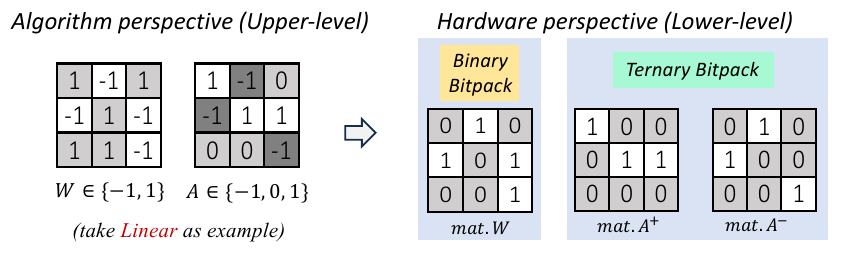}
    \caption{The difference of the algorithm (upper-level) perspective and the hardware (lower-level) perspective in BWTA quantization scheme. }
    \label{fig:low_level_perspective}
    \vspace{-0.6em}
\end{figure}

There are three major arithmetic rules in Transformer-based models: 
(1) Linear layers, where the weight is signed $(-1, 1)$, while the ternary activation are $(-1, 0, 1)$. 
(2) Attention layers, where the attention matrix is $\{0, 1\}$ by $\operatorname{bool}$, and the value matrix is $\{-1, 0, 1\}$. 
(3) MatMul layers, where the query and key matrices are both quantized to $\{-1, 0, 1\}$. 

\begin{figure}
 \vspace{-1.0em}
    \centering
    \includegraphics[width=0.94\textwidth]{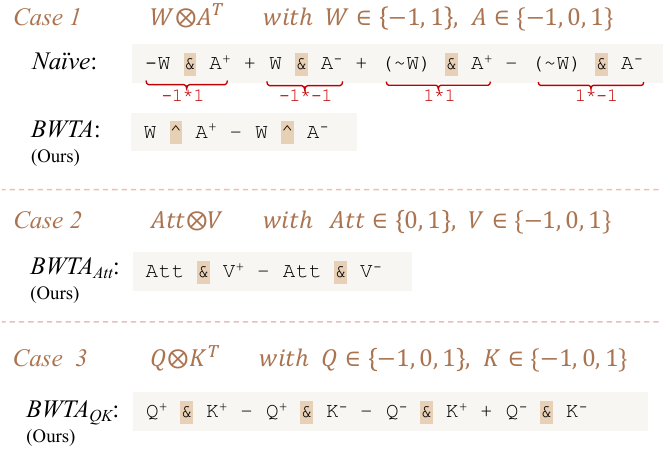}
    \caption{The arithmetic rules of the three layers.  {\texttt{\^}} and {\texttt{\&}} means \texttt{mma} instructions with \texttt{xor} and \texttt{mma} instructions with \texttt{and}, respectively. We dub them as \texttt{mma.and} and \texttt{mma.xor} in the main text. }
    \label{fig:formula_of_three_cases}
     \vspace{-0.6em}
\end{figure}

The underlying binarized \texttt{mma} instructions consist of binary connective logic gates (\texttt{and} or \texttt{xor}) for bitwise multiplication and a \texttt{popcount} for accumulation. As simple yet versatile logic, it allows us to design various arithmetic rules through combinations of instructions to support the upper-level algorithmic rules. 
Fig.~\ref{fig:formula_of_three_cases} shows three cases of matrix multiplication in Transformer-based architectures. {\texttt{\^}} and {\texttt{\&}} denote \texttt{mma} instructions with \texttt{xor} or \texttt{and}. In Binary Bitpack, positive numbers are stored as bit $0$, while negative numbers are stored as bit $1$. In Ternary Bitpack, we use two registers to store positive and negative values respectively (denoted as $^+$ and $^-$). 
We introduce each variant in detail. 

\modified{\textit{Case 1}: $W\otimes A^\top$ for Linear layers. It represents the standard MatMul where weights are binarized to $\{-1, 1\}$ and the activations are quantized to $\{-1, 0, 1\}$. 
Straightforwardly, the calculation process can be \texttt{-W \& A$^+$ + W\&A$^-$ + ($\sim$W) \& A$^+$ - ($\sim$W \& A$^-$)}, which involves four \texttt{mma.and}, one \texttt{not} and four \texttt{add/sub}. Fortunately, it can be simplified to two \texttt{mma.xor} and one \texttt{sub} by equivalent substitution. As shown in Fig.~\ref{fig:formula_of_three_cases} \textit{Case 1}, we name it {\texttt{xor}-based BWTA MatMul}. It is a faster and more streamlined design, directly reducing the number of instructions by half. }

\modified{\textit{Case 2}: $Att\otimes V$ for MatMul of attention scores and value matrix. 
In this case, we have $Att\in\{0, 1\}$ and $V\in\{-1, 0, 1\}$. As Fig.~\ref{fig:formula_of_three_cases} \textit{Case 2} shows, it requires two \texttt{mma.and} and one \texttt{sub} instructions to compute the $1$ and $-1$ in $V$ separately. }

\modified{\textit{Case 3}: $Q\otimes K^\top$ for MatMul of query and key matrices. Both matrices are represented as $\{-1, 0, 1\}$. As shown in Fig.~\ref{fig:formula_of_three_cases} \textit{Case 3}, we need four \texttt{mma.and} and four \texttt{and/sub} to compute the results. 
Meanwhile, it requires two additional registers to store the intermediate results for accumulation. }

By defining the above three algorithmic rules with instruction combinations, it is able to implement BWTA in most Transformers. We separately test the three BWTA variants in the ablation study (Sec.~\ref{sec:exp:efficiency}) to verify the time consumption of bitpacking and multiplication for the designed kernels, as well as its end-to-end accleration benefits. 

%% file: sec/4_experiments.tex
\section{Experiments}  
\label{sec:experiments}

In this section, we conduct extensive experiments to validate the performance of our proposed BWTA quantization framework, focusing on both the accuracy and efficiency. 
We begin with an ablation study for Smooth Multi-Stage Quantization in Sec.~\ref{sec:exp:ablation}, with a detailed analysis of the parameter convergence and training stability in Sec.~\ref{sec:exp:convergence}. 
Next, we assess the model accuracy on both BERT and LLM-based models, comparing with the SOTA low-bit methods in Sec.~\ref{sec:exp:accuracy}. 
In final, we provide a comprehensive evaluation of our BWTA CUDA kernel, detailing the efficiency for both kernel-level and end-to-end performance in Sec.~\ref{sec:exp:efficiency}.

\subsection{Implementation Details}
\label{sec:implementation_details}

\textbf{Datasets.} 
The evaluation on BERT-based models are conducted on GLUE benchmark~\cite{wang2018glue}, consisting of nine basic language tasks. 
We exclude WNLI task as previous studies do for its relatively small data volume and unstable behavior. 
\new{We evaluate LLM-based models on Wikitext2 and C4 datasets to compare the perplexity, and accuracy performance on CommonsenseQA benchmarks. 
}

\textbf{Models. }
For BERT-based, we follow~\cite{bai2021binarybert} to take well-trained DynaBERT~\cite{hou2020dynabert} as the teacher model to self-supervise the training of the binarized ones. \new{For LLM-based, we take the Bitnet series with sizes of 700M, 1.3B and 3B as the pretrained weights. }
For all experiments, we binarize the weight in linear layers by $\operatorname{sign}$, apply $\operatorname{bool}$ for attention probability and activation function, and $\operatorname{ternary}$ for all the other activations. Details can be found in Appendix.~\ref{sec:baseline_scheme}. 

\textbf{Baselines}. 
For BERT-based, we compare with previous quantization methods on BERT, including Q2BERT~\cite{zafrir2019q8bert}, TernaryBERT~\cite{zhang2020ternarybert}, BinaryBERT~\cite{bai2021binarybert}, BiBERT~\cite{qin2022bibert}, BiT~\cite{liu2022bit}, and more recent works such as BEBERT~\cite{bebert}, BiPFT~\cite{xing2024bipft}, MLBERT~\cite{mlbert}. 
\new{For LLM-based, we compare with RTN (round-to-nearest), GPTQ~\cite{frantar2022gptq}, and 1-bit methods PB-LLM~\cite{shang2023pbllm}, BiLLM~\cite{huang2024billm}, Bitnet~\cite{wang2023bitnet}, and reproduce BWN~\cite{liu2022bit} on LLMs. }

\subsection{Ablation Study}
\label{sec:exp:ablation}

\subsubsection{Comparison of Multi-stage Quantization Strategies}
\label{sec:ablation_multistage}

\input{figs/fig_ablation_multi_stage}

In Fig.~\ref{exp:fig:ablation_levelwise}, we compare the levelwise and bitwise degradation strategies for multi-stage quantization. The X-axis is the training epochs. The {yellow} line represents the bitwise multi-stage quantization, starting with $2^3$ integers. The {blue} line denotes the levelwise multi-stage quantization, which begins at $L_0=4$ decreases by 1. Both reinitializing the scaling factors at each transition. The black line shows the levelwise strategy in addition to the proposed magnitude alignment projection factor. 

Compared to the bitwise multi-stage quantization, the levelwise approach exhibits a more gentle (or no significant) accuracy drop as the number of levels decreases, recovering more quickly within a few epochs during stage transition. Similarly, the loss shows a smaller increase when reducing to fewer levels, reflecting a smoother stage transition. 

\subsubsection{Effect of Projecting Factor}
\label{sec:ablation_factor}
\input{figs/fig_ablation_factor}

\modified{We compare the Magnitude Alignment Projection Factor with with three alternative scaling factor re-initialization methods at the stage transitions: \textbf{Search}, performing layerwise searching following \cite{yuan2021ptq4vit} ({grey}); \textbf{Mean}, initializing with the mean of absolute values following \cite{liu2022bit} ({green}); \textbf{None}, directly using the factor from the previous stage without adjustment ({red} dashed line).}

Fig.~\ref{exp:fig:ablation_factor} illustrates the losses and change of scaling factors during training.
The results indicate that layerwise searching (\textbf{Search}) consistently yields poor initialization, requiring several epochs to recover and find the optimal scaling range. It suggests that while the searching strategy may minimize quantization error at the layer level, it is suboptimal when considering the overall performance of the network. 
Similarly, the \textbf{Mean} approach is also inadequate. Although it reflects the magnitude of the original values, it gives up the previously learned scaling factors, which have already been adapted to the current model parameters. However, applying the projection factor (\textbf{Ours}) provides a more effective correction for the scaling factors. It better aligns the magnitude of the parameters between adjacent stages, makes the scaling factors start from an initial value closer to the optimal. The loss curves indicate that, using the projection factor not only achieve lower losses at stage transition, but also converges faster afterward, serving as an effective solution for stable training. 

\input{tables/accuracy_glue_w_DA}

\subsection{Accuracy Comparisons to SOTA methods}

\label{sec:exp:accuracy}

On \textbf{BERT-based} models, we compare the binarized BWTA BERT with other binarization methods on GLUE benchmark, including recent works BEBERT~\cite{}, BiPFT~\cite{} and MLBERT~\cite{}, along with the average accuracy across eight tasks displayed in the right column in TABLE~\ref{tab:glue-w-DA}. 
We also list the model size of each method, where our BERT has the same storage compared to other binarization methods due to its 1-bit weight. 
It shows that our BWTA BERT surpasses the previous SOTA method by an obvious margin, which is up to 3\% on average while bringing no storage overheads. The accuracy performance is even approaching the full-precision counterpart in most of the tasks, such as QQP, QNLI, SST-2, STS-B, MRPC and RTE, which are only about 2\% performance drops. 
Meanwhile, our BWTA also surpasses most of the BERT models with higher bitwidth at the same model sizes, such as W2A8, W1A4, showing great potential for extreme quantization. 

\input{tables/accuracy_ppl}

On \textbf{LLM-based} models, we compare the perplexity on WikiText-2 and C4, and the average accuracy on CommonsenseQA benchmarks in TABLE~\ref{tab:accuracy-llm-ppl}. (The detailed scores in each task of commonsenseQA can be found in TABLE~\ref{tab:accuracy-llm-cmQA} in Appendix~\ref{app:exp_llm}.) Results show that our BWTA achieves comparable accuracy even with lower average bitwidth and model sizes, indicating negligible degradation in generation quality. 
As the first work to quantize activations to such a low bitwidth, our BWTA shows great potential for extreme LLM quantization. 

\subsection{Efficiency Experiments}
\label{sec:exp:efficiency}

We comprehensively verify the efficiency of our BWTA kernel by providing kernel-level time breakdown of each component individually as well as the end-to-end performance in both prefill/decode phases on LLMs.

\input{tables/speed_kernel_benchmark}

\subsubsection{Kernel-Level Speedup}


\new{We conduct kernel-level benchmarks comparing to other acceleration linear kernels, including: (1) \texttt{torch.nn.functional} (FP16/BF16 linear); (2) \texttt{bnb.nn.Linear4bit} (bitsandbytes 4-bit weight-only quantization)\footnote{https://huggingface.co/docs/bitsandbytes/en/reference/nn/linear4bit}; (3) \texttt{bitlinear\_int8xint2} (BitNet ternary-weight / int8-activation)\footnote{https://github.com/microsoft/BitBLAS}. }

\new{As shown in TABLE~\ref{tab:kernel_bench_256}, our kernel provide five matrix-multiply variants for binary/ternary models, which can be used as a plug-in low-level operator for other binary/ternary methods. All custom kernels are implemented in C++ and  compiled as CPython extension modules. }


\new{Results show that under the same GPU/driver/CUDA stack, our ultra low-bit kernels outperform all the competitives. Moreover, ultra-low-bit kernels yield higher throughput than higher-bit linear baselines; notably, our best configuration exceeds the BitNet int8$\times$int2 kernels ($5.545\times10^3$ FLOPs vs. $4.068\times10^3$ FLOPs). 
More kernel-level benchmark results can be found in the Appendix~\ref{app:more_results_of_kernel_benchmark}. }

\subsubsection{Latency Breakdown}

\begin{figure*}
    \centering
    \includegraphics[width=0.932\linewidth]{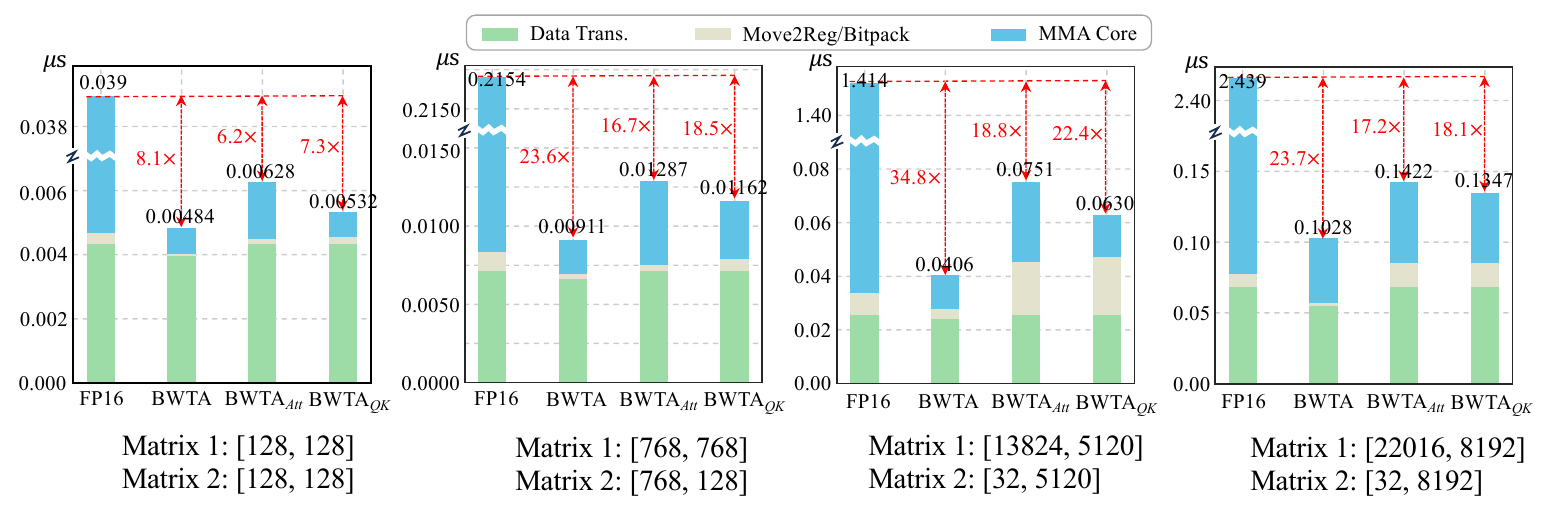}
    \caption{Overall time comparisons for each step in the GEMM kernels with four typical shapes. }
    \label{fig:ablation_latency_step}
    \vspace{-1.0em}
\end{figure*}

Fig.~\ref{fig:ablation_latency_step} show the time consumption of each step in different GEMM kernels with four shapes, ranging from small matrices shapes commonly used in BERT, to larger matrices shapes in LLMs (such as $13824\times 5120$ and $22016 \times 8192$ in Llama-7b). The time spent in each step is depicted in bars. 
The speedup ratios of each BWTA MatMul kernel variant compared to their FP16 counterparts are indicated by red dashed lines. 
The whole process is divided into three parts: 

(1) \textit{Data Trans.} ({green} bar) refers to the data transmission step. 
This stage also includes casting the \texttt{mma} results from \texttt{INT32} to \texttt{half} before offloading them to memory. From the histogram, we can see that BWTA is slightly faster than the FP16 baseline in data transmission. 
The speed advantage is especially notable when the weight matrix is large, with a time reduction from 0.068$\mu$s for FP16 to 0.054$\mu$s for BWTA in a single MatMul operation with shapes of $22016\times 8192$ and $32\times 8192$. 

(2) \textit{Move2Reg/Bitpack} ({cream} bar) represents moving or packing data from shared memory to registers. 
We can see that the \textit{Bitpack} does not spend much extra time compared to \textit{Move2Reg}, regardless of the matrix size. 

(3) \textit{MMA Core} ({blue} bar) is the matrix multiplication step. 
As the blue bars in Fig.~\ref{fig:ablation_latency_step} show, the FP16 MMA core with \texttt{m16n8k16} layout costs 20-40$\times$ time more than the BWTA MMA core with \texttt{m8n8k128}. And the ratio can even reach up to 50$\times$ in larger matrix shapes. 
The acceleration can be attributed to two aspects: the binarized \texttt{mma} instruction has inherently lower latency, and accommodates larger shapes than the FP16 one, which means higher parallelism, higher throughput, and fewer execution cycles.  

Overall, the BWTA MatMul Kernel significantly outperforms the FP16 counterpart in execution speed especially at larger matrix shapes. 
The detailed ablation study on kernel designs can be found in Appendix~\ref{app:ablation-kernel-speed}

\input{tables/e2e_speed}

\new{\subsubsection{End-to-End Speedup}
We include three mainstream quantization frameworks/toolkits with comparable model sizes: (1) Llama-3.2-3B implemented with \textit{bitsandbytes} (\textbf{4-bit}, QLoRA; NF4)\footnote{https://huggingface.co/docs/bitsandbytes/en/reference/nn/linear4bit}, (2) Gemma-2B with BF16\footnote{https://huggingface.co/google/gemma-2b} and quantization via \textit{AutoGPTQ}\footnote{https://huggingface.co/Intel/gemma-2b-int4-inc} (\textbf{4-bit}, weight-only). (3) Bitnet-b1.58-2B with customized low-bit matmul kernel (int8$\times$int2 linear operator) provided by \textit{Microsoft}\footnote{https://github.com/microsoft/BitNet/tree/main/gpu}. }
\new{We report prefill and decode jointly in TABLE~\ref{tab:prefill-bench-decode}. Time in these tables is the phase-only latency (i.e., not prefill+decode combined). Results show that: }

\new{\textbf{Prefill}. We present latency (s) and throughput (tokens/s) across batch sizes \{1, 4, 8, 16\} at 2k context length. 
The acceleration gains are especially clear at larger batch sizes: for batch=16, replacing 10\%/15\% of the parameters with BWTA yields approximately 216/330 tokens/s additional throughput, respectively. It shows an approximately linear speedup as more layers and parameters are replaced to BWTA. GPU memory usage is also reduced. Notably, the weight-only 4-bit methods (Llama-3.2-3B with bnb-4bit QLoRA and Gemma-2B with 4-bit AutoGPTQ) run out of memory at batch $\ge8$ with much less throughput. 
}

\new{\textbf{Decode}. We report generation time (s) and throughput (tokens/s) for lengths $\approx$50/100/150. Compared with W4A16 weight-only quantization, ultra low-bit quantization provides a significant latency reduction. BWTA achieves additional gains over BitNet-b1.58. When 15\% of layers are replaced with BWTA, we observe 12-15 tokens/s improvement in decoding throughput. }

\subsection{Visualizations and Analysis}
\label{sec:exp:convergence}

\subsubsection{Activation Visualization}
\label{sec:visualization_activation_distribution}

We visualize the quantized activations in Fig.~\ref{fig:visualization_activation}. It shows the distribution of quantized activations along with training epochs for (a) the bitwise and (b) the levelwise degradation strategies. Deeper colors indicates a larger number of activations at that particular magnitude. 
This histogram helps us to clearly observe how the quantized activations evolve throughout the training process. 

\input{figs/fig_visualization_activation}

By comparing the two strategies, we find that in bitwise strategy, the quantization bins are not fully utilized due to its imbalance and mismatch with the original distribution. 
When transiting to the next stage with fewer integer levels, it causes a sudden shift in the activation distribution, resulting in an increase of loss and a drop in accuracy (refer to Fig.~\ref{exp:fig:ablation_levelwise}). 

On the contrary, the levelwise degradation strategy makes better use of the available quantization bins, creating a balanced distribution and retaining more information. Meanwhile, values close to zero can be consistently represented as zero, which enables a constant representation and stabilizes the training. 
In conjunction with the projection factor, the magnitudes of the activation remain stable at stage transition, allowing for a smooth convergence. 


\subsubsection{Loss Surface Visualization}

We visualize the surface of the task loss to evaluate the robustness of the binarized models in Fig.~\ref{fig:loss_surface}. It illustrates the loss surface of (a) full-precision model and (b) the BiT and (c) the BWTA models. 
To analysis the impact of perturbations, we apply a minor shift $\theta \in [-0.2, 0.2]$ to the trained weights, represented on $x$ and $y$ axes as two perturbation directions. 

\input{figs/fig_loss_surface}

The visualization reveals two findings: 
(1) \textit{Lower task loss}: BWTA model achieves a lower loss than BiT (1.05 vs. 1.62), indicating that our BWTA has converged to a lower point, delivering a better accuracy performance. 
(2) \textit{Lower Deviation}: The BWTA model exhibits smaller standard deviation of the loss, which is 0.026 lower than that of BiT. It means that the BWTA model is more robust to perturbations. 
In a word, compared to other binarized methods, our BWTA model reaches a better optimum with the proposed bitwidth scheme and the levelwise training framework. It demonstrates the effectiveness of our approach in improving the convergence and robustness of ultra-low bitwidth Transformers.

%% file: figs/fig_ablation_multi_stage.tex
\begin{figure}[t]
    \centering
    \includegraphics[width=0.8\linewidth]{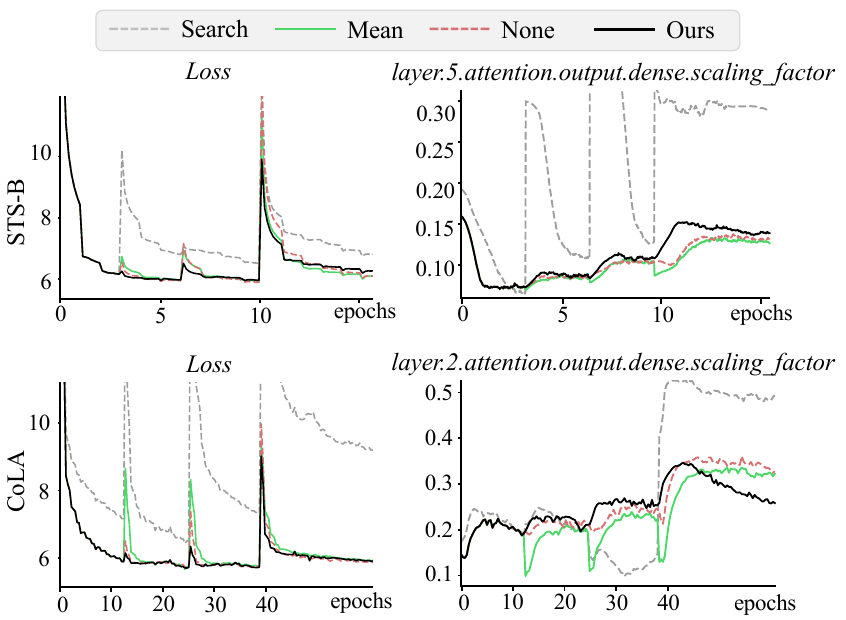}
    \caption{Comparison of level/bitwise multi-stage quantization strategies (start with $L_0=4$ and $k_0=3$ as an example). 
    }
    \label{exp:fig:ablation_levelwise}
\end{figure}

%% file: figs/fig_ablation_factor.tex
\begin{figure}[t]
    \centering
    \includegraphics[width=0.8\linewidth]{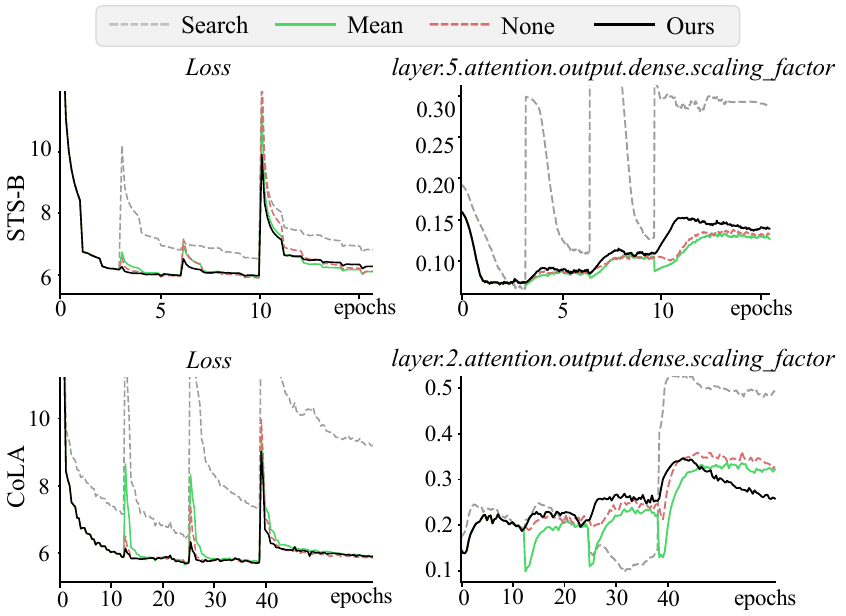}
    \caption{Comparison of different re-initialization strategies for scaling factor. }
    \label{exp:fig:ablation_factor}
\end{figure}

%% file: tables/accuracy_glue_w_DA.tex
\begin{table*}[htb]
    \caption{Comparison of BERT quantization methods. All results are trained with data augmentation except BiPFT. ``--'' means the results are not officially reported. Experiments without data augmentation are provided in Appendix~\ref{app:glue-wo-DA}. *Data augmentation is not needed for MNLI and QQP.}
    \label{tab:glue-w-DA}
    {\footnotesize
    \centering
    \resizebox{0.98\textwidth}{!}{
    \begin{tabular}{llccccccccccc}
    \toprule
    \textbf{Quant}  & \begin{tabular}[c]{@{}c@{}}\textbf{\#Bits}\end{tabular} & \begin{tabular}[c]{@{}c@{}}\textbf{Size}$_\text{ (MB)}$\end{tabular}  & \begin{tabular}[c]{@{}c@{}}\textbf{MNLI}$_\text{-m/mm}$\end{tabular} & \textbf{QQP} & \textbf{QNLI} & \textbf{SST-2} & \textbf{CoLA} & \textbf{STS-B} & \textbf{MRPC} & \textbf{RTE} & \textbf{Avg.} \\ 
    \midrule
    Full Precision & 32/32 & 418 & 84.9/85.5 & 91.4 & 88.6 & 93.2 & 59.7 & 89.8 & 86.2 & 72.2  & 83.9 \\ 
    \midrule
    Full Precision & 32/32 & 418 & 84.9/85.5* & 91.4* & 92.1 & 93.2 & 59.7 & 90.1 & 86.3 & 72.2  & 82.3 \\
    {TernaryBERT} & 2/8 &  28.0 & 83.3/83.3* & 90.1* & 90.0 & 92.9 & 47.8 & 84.3 & 82.6 & 68.4 & 77.8\\
    {TernaryBERT} & 2/2 & 28.0 & 40.3/40.0* & 63.1* & 50.0 & 87.5 & 20.6 & 72.5 & 72.0 & 47.2 & 58.3 \\
    {TernaryBERT} & 2/1 & 28.0 & 32.7/33.0* & 74.1* & 50.9 & 80.3 & 6.5 & 10.3 & 71.5 & 53.4 & 45.5 \\
    {BinaryBERT} & 1/4 & 16.5 & 83.9/84.2* & 91.2* & 91.4 & 93.7 & 53.3 & 88.6 & 86.0 & 71.5 & 80.8\\
    \new{BEBERT} &  1/4 &  33.0 &  84.7/84.7* &  91.4* &  91.5 &  93.4 & 56.5  &  -- &  85.2 &  71.5  &  82.4 \\ 
    \new{MLBERT} &  1/4 & --  &  -- & --  &  -- & 93.5  & 53.5 & -- &  86.1 &  72.2 &  76.3 \\
    \new{MLBERT} &  1/3 & 13.0  & -- & --  & -- & 93.4  &  52.1 & -- &  85.8 &  69.2 & 75.1 \\
    BinaryBERT & 1/2 & 16.5  & 62.7/63.9* & 79.9* &  51.0 & 89.6 & 33.0 & 11.4 & 71.0 & 55.9 & 52.0 \\
    \new{BiPFT} &  1/2 &  14.9 &  77.0/76.9* &  89.0* &  86.6 &  88.1 & 36.2 &  87.0 &  84.1 &  66.1 &  76.8 \\
    \new{MLBERT} & 1/2 & --  &  -- & --  &  --  & 92.6 &  50.2 & -- & 85.2 &  68.1 & 74.0 \\
    \midrule
    BinaryBERT & 1/1 & 16.5 & 35.6/35.3*  & 66.2* & 66.1 & 78.3 & 7.3 & 22.1 & 69.3 & 57.7 & 50.1 \\
    BiBERT &  1/1 & 13.4 & 66.1/67.5* & 84.8* & 76.0 & 90.9 & 37.8 & 56.7 & 78.8 & 61.0 & 67.0 \\ 
    BiT &  1/1  & 13.4 & 79.5/79.4* & 85.4* & 82.5  & 92.3  & 38.2  & 84.2  & {83.0}  & 69.7 & 77.5 \\ 
    \new{BEBERT} &  1/1 & --  &  75.7/76.6*  &  84.9* &  80.7 &  91.8 & 42.6 & -- &  82.4 &  65.8 & 74.7 \\ 
    \textbf{Ours} & \textbf{1/1.5} & \textbf{13.4} & \textbf{81.0/81.3*} & \textbf{89.2*} & \textbf{85.5} & \textbf{92.9} & \textbf{49.3} & \textbf{87.4} & \textbf{85.5} & \textbf{72.2} & \textbf{80.4} \\
    \bottomrule
    \end{tabular}
    }}
    \end{table*}

%% file: tables/accuracy_ppl.tex
\begin{table}[ht]
    \centering
    \caption{\new{Perplexity on WikiText2 and C4, and average accuracy on CommensenseQA for various models and quantization methods.  ``--'' means the result is not officially reported. For all BWN and BWTA results, we replace 30\% layers to ultra-low bitwidth based on Bitnet~\cite{wang2023bitnet}.}}
    \label{tab:accuracy-llm-ppl}
    \begin{newtable}
    \resizebox{\textwidth}{!}{
    \begin{tabular}{l l c r r r}
    \toprule
    \textbf{Model} & \textbf{Method} & \textbf{W$_{bit}$/A$_{bit}$} & \textbf{Wikitext2} & \textbf{C4} & \textbf{CmQA} \\
    \midrule
    \multirow{4}{*}{LLaMA-2-7B}
      & Full Precision & 16/16     & 5.47      & --  &  68.9     \\
      & GPTQ           & 2/16      & 20.85     & --  & 37.5     \\
      & PB-LLM         & 2/16      & 20.37     & --  & 40.0     \\
      & BiLLM          & 1.1/16   & 32.48     & --   & 42.3   \\
    \midrule
    \multirow{4}{*}{OPT-2.7B}
      & RTN            & 2/16      & 9505.76   & --  & --    \\
      & GPTQ           & 2/16      & 61.59     & --   & --   \\
      & PB-LLM & 1.7/16   & 124.35    & --  & --   \\
      & BiLLM          & 1.1/16   & 49.55     & --  & --   \\
    \midrule
    \multirow{5}{*}{OPT-1.3B}
      & RTN            & 4/16      & --        & 20.25  & --   \\
      & RTN            & 2/16      & 11272.65  & --  & --   \\
      & GPTQ           & 2/16      & 115.17    & --  & --   \\
      & PB-LLM & 1.7/16   & 265.52    & 17.60  & --   \\
      & BiLLM          & 1.1/16   & 69.97     & --  & --   \\
    \midrule
    \multirow{3}{*}{Bitnet-b1.58-1.3B}
      & Bitnet         & 1.5/8     & 11.20   & 11.17  & 47.2 \\
      & BWN~\cite{liu2022bit}        & 1.3/6     & 20.57  & 22.31 & 42.5\\
      & \textbf{BWTA (Ours)}    & \textbf{1.3/6} & \textbf{15.58}   & \textbf{16.09} & \textbf{44.8} \\
    \midrule
    
    \multirow{3}{*}{Bitnet-b1.58-0.7B}
      & Bitnet         & 1.5/8     & 13.36    & 12.20  & 44.4 \\
      & BWN~\cite{liu2022bit}        & 1.3/6     & 21.59  & 29.85 & 40.4 \\
      & \textbf{BWTA (Ours)}    & \textbf{1.3/6} & \textbf{19.04}  & \textbf{21.04} & \textbf{43.4} \\
    \bottomrule
    \end{tabular}}
\end{newtable}
\end{table}

%% file: tables/speed_kernel_benchmark.tex

\begin{table}[h]
\centering
\caption{\new{Kernel-level benchmark. All test cases are repeated 50 times with 256$\times$256 multiplying with 256$\times$1 on NVIDIA H800 GPU (96GB VRAM).} }
\resizebox{\textwidth}{!}{
\new{\begin{tabular}{l c r r}
\toprule
\textbf{Kernel} & \textbf{W$_{bit}$/A$_{bit}$} & \textbf{Time ($\mu$s)} & \textbf{FLOPs ($\times 10^3$)} \\
\midrule
\texttt{torch.nn.functional }           & 16/16   & 10.74  & 3.05 \\
\texttt{bnb.nn.Linear4bit}              & 4/16    & 60.33 & 0.543 \\
\texttt{bitlinear\_int8xint2}(Bitnet)  & 2/8     & 9.08* & 3.609 \\  \midrule
Binary Linear   (Supported)        & 1/1     & 7.29  & 4.495 \\
Binary A x V   (Supported)          & 1/1     & 6.97  & 4.701 \\
BWTA (Ours)               & 1/1.5   & 8.56  & 3.828 \\
BWTA$_{Attn}$ (Ours)      & 1/1.5   & 6.71  & 4.883 \\
BWTA$_{QK}$ (Ours)         & 1.5/1.5 & 5.54  & 5.915 \\
\bottomrule
\end{tabular}
\label{tab:kernel_bench_256}
}}
\end{table}

%% file: tables/e2e_speed.tex
\begin{table*}[t]
\centering
\caption{\new{End-to-end benchmarks on various models with $\le$3B parameters. The input token length of prefill phase is uniformly set to $2k$, and the batch size of decode phase is 1. ``G. Len.'' means the generated token length during decoding.} }
\label{tab:prefill-bench-decode}
\resizebox{0.95\textwidth}{!}{
\begin{newtable}
\begin{tabular}{l cc | crrr | rrr}
\toprule
\multirow{2}{*}{\textbf{Models} } &  \multirow{2}{*}{\textbf{W$_{bit}$/A$_{bit}$}} & \multirow{2}{*}{\textbf{Storage (GB)}} & \multicolumn{4}{c|}{\textbf{Prefill Phase (SeqLen=2k)}} & \multicolumn{3}{c}{\textbf{Decode Phase (BS=1)}}  \\ 
 & & & \textbf{BS} & \textbf{Time (s)} & \textbf{Tokens/s} & \textbf{Mem. (GB)} & \textbf{G. Len.} & \textbf{Time (s)} & \textbf{Tokens/s} \\
\midrule
\multirow{3}{*}{Llama-3.2-3B}   & \multirow{3}{*}{4/16 (bnb)} &  \multirow{3}{*}{2.240}  & 1 & 0.072 & 28267 & 15.59    & 50 & 1.317 & 43.4 \\
&  & & 4 & 0.172 & 11924 & 54.94 & 100  & 2.563 & 39.0  \\
&  & & 8 & --      & --    & OOM   & 150  & 3.507 & 44.2    \\  \midrule
\multirow{6}{*}{Gemma-2B}     & \multirow{3}{*}{16/16}  &  \multirow{3}{*}{5.012} & 1 & 0.063 & 32492 & 20.75 & 50 & 0.512 & 111.6   \\
&    & & 4 & 0.116 & 17694 & 64.51  & 100  & 1.146 & 87.3 \\
&    & & 8 & --      & --          & OOM    & 150  & 1.860 & 83.3    \\  \cmidrule{2-10}
& \multirow{3}{*}{4/16 \makecell[c]{(gptq)}}  & \multirow{3}{*}{3.130}  & 1  & 0.107 & 19133 & 22.53  & 50 & 0.989 & 50.8  \\
& & & 4  & 0.173 & 11858 & 57.71  & 100  & 1.959 & 51.0 \\
& & & 8  & --      & --          & OOM    & 150  & 2.932 & 52.9    \\ \midrule
\multirow{9}{*}{Bitnet-b1.58-2B} & \multirow{3}{*}{1.5/8}  &  \multirow{3}{*}{1.835} & 1  & 0.026 & 77253  & 10.56  & 50 & 0.158 & 360.2 \\
&  & & 8  & 0.189 & 10814  & 31.67  & 100  & 0.295 & 331.0 \\
&  & & 16 & 0.378 & 5415  & 55.60 & 150 & 0.458 & 333.9  \\  \cmidrule{2-10}

& \multirow{3}{*}{\makecell[c]{1.45/7.35 \\ (10\% layers)}}   & \multirow{3}{*}{\textbf{1.819}} 
&  1  & \textbf{0.026} & \textbf{77225} & \textbf{10.28}  & 50   & \textbf{0.156} & \textbf{365.9} \\
&  & & 8  & \textbf{0.185} & \textbf{11096} & \textbf{31.48} & 100   & \textbf{0.290} & \textbf{338.0}  \\
&  & & 16 & \textbf{0.369} &  \textbf{5544} & \textbf{55.41}  & 150  & \textbf{0.449} & \textbf{338.4}\\ \cmidrule{2-10} 
 
& \multirow{3}{*}{\makecell[c]{1.35/6.05 \\ (30\% layers)}} & \multirow{3}{*}{\textbf{1.796}} 
 & 1  & \textbf{0.025} & \textbf{81529} & \textbf{10.10}  & 50   & \textbf{0.152} & \textbf{375.4}  \\
& & & 8  & \textbf{0.179} & \textbf{11466} & \textbf{32.02}  & 100   & \textbf{0.285} & \textbf{344.4}  \\
& & & 16 & \textbf{0.356} &  \textbf{5745} & \textbf{56.94}  & 150  & \textbf{0.443} & \textbf{345.8} \\
\bottomrule
\end{tabular}
\end{newtable}
}
\end{table*}

%% file: figs/fig_visualization_activation.tex
\begin{figure}[t]
        \centering
        \includegraphics[width=\linewidth]{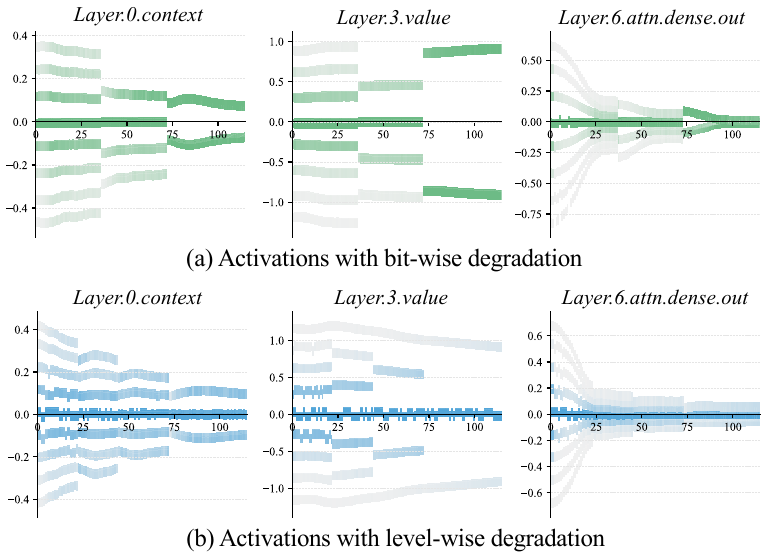}
        \caption{The quantized activation of (a) bitwise and (b) levelwise degradation. }
        \label{fig:visualization_activation}
    \vspace{-1em}
\end{figure}

%% file: figs/fig_loss_surface.tex
\begin{figure}[ht]
    \centering
    \includegraphics[width=\linewidth]{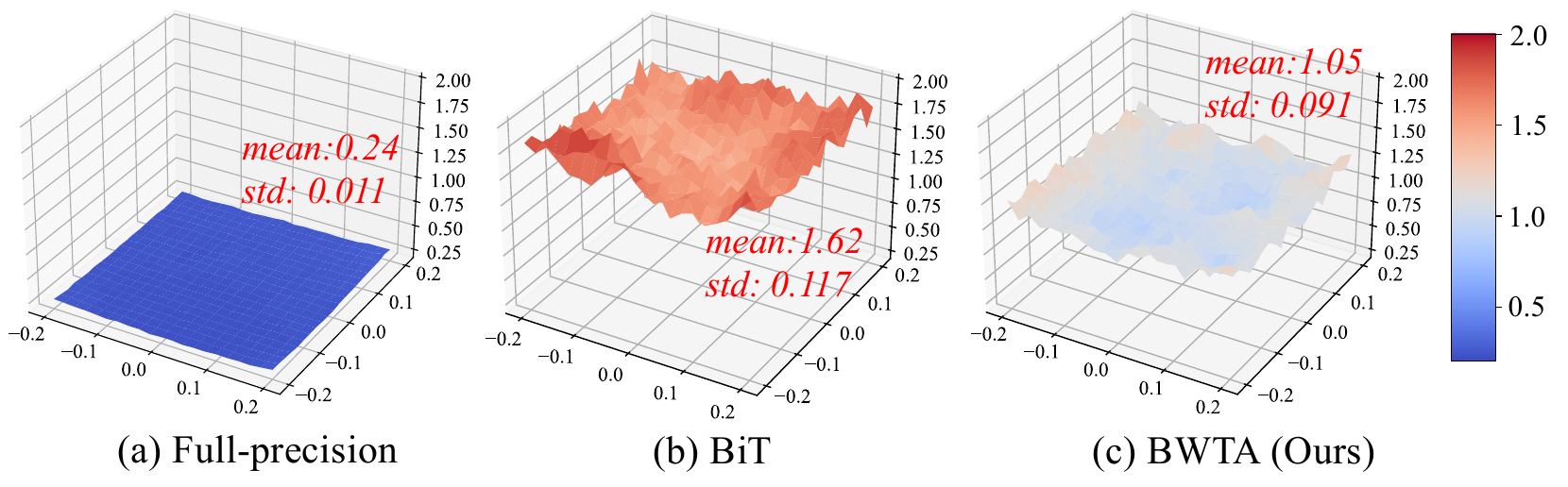}
    \caption{Task loss surface of quantized models and the full-precision counterpart. }
    \label{fig:loss_surface}
    \vspace{-0.1in}
\end{figure}

%% file: sec/5_conclusion.tex
\section{Conclusion}

We propose the Binary Weight \& Ternary Activation (BWTA) quantization framework to address the issues of zeropoint distortion, and improve the accuracy of ultra-low bitwidth Transformer-based models through algorithm-hardware co-design. We effectively train the BWTA models by Smooth Multi-stage Quantization, stabilizing the convergence in ultra-low bitwidth representation. Meanwhile, we design the first full-stack GPU MatMul kernels to ensure the real acceleration of BWTA ops. BWTA not only achieves state-of-the-art accuracy and efficiency, but is also versatile to be applied to various Transformer-based models, benefiting from significant speedups on GPUs. This work paves the way for further studies for ultra-low bitwidth Transformers, demonstrating its practical value through both good accuracy performance and low inference latency.

%% file: sec/appendix/7_appendix.tex
\appendices

\section{More Rationales of Our Motivation}
\input{sec/appendix/kernel_profile}

\input{sec/3_baseline}

\section{Additional Experiments}

\input{sec/appendix/exp_details}

\input{sec/appendix/gradient_of_scaling}

\input{sec/appendix/exp_llm}

\input{sec/appendix/more_results_of_kernel_benchmark}

\input{sec/appendix/limit}

%% file: sec/appendix/kernel_profile.tex
\input{tables/speed_profile}

\new{\subsection{Operator-Level Latency Breakdown}
In our study, we target the dominant GPU bottlenecks (i.e., matrix multiplications in linear/attention) and keep nonlinearities in higher precision (BF16 for LLMs; FP32 for BERT). This choice is motivated by the following rationales: 
\begin{itemize}
    \item \textbf{GEMM dominate latency. } We report an operator-wise breakdown in Table~\ref{tab:profile-prefill} and Table~\ref{tab:profile-decode} for prefill and decode phases. It shows the total elapsed device-side CUDA time for different operator categories. 
    Results demonstrate that GEMM and attention pathways cost the most end-to-end latency on GPUs, while Softmax/LayerNorm/GELU contribute a comparatively small share. For Llama, when dealing with longer sequence length ($>$1k) and larger batch sizes (4/8), and for all tested cases on BERT, time spent on matrix multiplications are always much more than the nonlinearities. 
    \item \textbf{Numerical stability.} Retaining LayerNorm/Softmax in higher precision stabilizes training and inference under ultra low-bit quantization, preventing degradation or collapse. This observation is reported and used in BitNet-b1.58~\cite{wang2023bitnet}, and is consistent with our findings. Meanwhile, many prior work on both BERT-based and LLM-based binarization methods also adopt the same practice to keep the nonlinearities with higher bitwidth floating-point operations~\cite{liu2022bit, qin2022bibert, huang2024billm, shang2023pbllm}. 
    \item \textbf{Integer quantization on nonlinearities may require customized hardware.} We acknowledge concurrent progress on quantizing nonlinearities with specialized hardware. Notably, SoftmAP~\cite{rakka2024} proposes an customized hardware, associative-processor (AP) architecture, which is built on content-addressable memory (CAM) to accelerate an integer-approximate softmax. SOLE~\cite{sole} designs E2Softmax and AILayerNorm as low-bit quantized variants of nonlinear functions, and implements by customized hardware unit on AISCs. 
    These work are promising for specialized hardware, while we adopted modern GPUs to ensure fair and reproducible comparisons with other low-bit methods, avoiding confounds from heterogeneous hardware resources and implementation. 
\end{itemize}
In sum, due to the lower time cost and representation stability of nonlinear operators on GPUs, we keep global nonlinear ops high-precision.}

%% file: tables/speed_profile.tex
\begin{table*}[h]
\centering
\caption{Operator-wise latency breakdown during prefill phase (device-side CUDA time). Results report the total elapsed time for each operator category. }
\resizebox{\textwidth}{!}{
\begin{tabular}{l c c
                r r
                r r
                r r
                r r}
\toprule
\multirow{2}{*}{\textbf{Model}} & \multirow{2}{*}{\textbf{BS}} & \multirow{2}{*}{\textbf{Seqlen}}
& \multicolumn{2}{c}{\textbf{\texttt{aten::*mm}}}
& \multicolumn{2}{c}{\textbf{\texttt{aten::softmax}}}
& \multicolumn{2}{c}{\textbf{\texttt{aten::*norm}}}
& \multicolumn{2}{c}{\textbf{\texttt{aten::*lu}}} \\
\cmidrule(lr){4-5}\cmidrule(lr){6-7}\cmidrule(lr){8-9}\cmidrule(lr){10-11}
 &  &  & \multicolumn{1}{c}{\textbf{Time (ms)}} &  \multicolumn{1}{c}{\textbf{Time (\%)}}  & \multicolumn{1}{c}{\textbf{Time (ms)}} &  \multicolumn{1}{c}{\textbf{Time (\%)}}  & \multicolumn{1}{c}{\textbf{Time (ms)}} &  \multicolumn{1}{c}{\textbf{Time (\%)}}  & \multicolumn{1}{c}{\textbf{Time (ms)}} &  \multicolumn{1}{c}{\textbf{Time (\%)}}  \\
\midrule
\multirow{9}{*}{Llama-3.2-3B} & \multirow{3}{*}{1} & 512  & 11.095  & 5.20\%  & 1.537   & 0.72\% & 6.393  & 3.00\% & 0.346 & 0.16\% \\
&  & 1024 & 21.057  & 11.42\% & 6.308   & 3.42\% & 5.961  & 3.23\% & 0.568 & 0.31\% \\
&  & 2048 & 44.245  & 9.55\%  & 25.092  & 5.42\% & 7.331  & 1.58\% & 1.178 & 0.25\% \\ \cmidrule(lr){3-11}
&  \multirow{3}{*}{4} & 512  & 38.365  & 14.21\% & 6.141   & 2.27\% & 7.387  & 2.74\% & 1.181 & 0.44\% \\
&  & 1024 & 78.264  & 12.43\% & 23.104  & 3.67\% & 13.196 & 2.10\% & 2.405 & 0.38\% \\
&  & 2048 & 173.133 & 9.76\%  & 97.943  & 5.52\% & 23.553 & 1.33\% & 4.839 & 0.27\% \\ \cmidrule(lr){3-11}
&  \multirow{3}{*}{8} & 512  & 74.146  & 14.51\% & 11.730  & 2.30\% & 13.072 & 2.56\% & 2.405 & 0.47\% \\
&  & 1024 & 155.640 & 12.70\% & 45.512  & 3.71\% & 23.509 & 1.92\% & 4.817 & 0.39\% \\
&  & 2048 & 347.385 & 9.85\%  & 194.505 & 5.51\% & 45.172 & 1.28\% & 9.737 & 0.28\% \\
\midrule
\multirow{9}{*}{bert-base}  & \multirow{3}{*}{1} & 128 & 5.687  & 15.83\% & 0.178 & 0.50\% & 0.317 & 0.88\% & 0.051 & 0.14\% \\
& & 256 & 6.866  & 20.32\% & 0.217 & 0.64\% & 0.335 & 0.99\% & 0.071 & 0.21\% \\
& & 512 & 10.520 & 24.93\% & 0.557 & 1.32\% & 0.359 & 0.85\% & 0.096 & 0.23\% \\ \cmidrule(lr){3-11}
&  \multirow{3}{*}{4}  & 128 & 9.548  & 24.42\% & 0.222 & 0.57\% & 0.356 & 0.91\% & 0.093 & 0.24\% \\
&  & 256 & 15.387 & 26.37\% & 0.487 & 0.83\% & 0.426 & 0.73\% & 0.165 & 0.28\% \\
&  & 512 & 26.139 & 25.08\% & 1.705 & 1.64\% & 0.693 & 0.67\% & 0.379 & 0.36\% \\ \cmidrule(lr){3-11}
&  \multirow{3}{*}{8}  & 128 & 14.798 & 27.59\% & 0.320 & 0.60\% & 0.426 & 0.79\% & 0.162 & 0.30\% \\
&  & 256 & 24.311 & 29.48\% & 0.864 & 1.05\% & 0.691 & 0.84\% & 0.387 & 0.47\% \\
&  & 512 & 49.772 & 29.65\% & 3.051 & 1.82\% & 1.065 & 0.63\% & 0.792 & 0.47\% \\
\bottomrule
\end{tabular}
\label{tab:profile-prefill}
}
\end{table*}

\begin{table*}[h]
\centering
\caption{Operator-wise latency breakdown during decode phase (device-side CUDA time). Results report the total elapsed time for each operator category.}
\resizebox{\textwidth}{!}{
\begin{tabular}{l c 
                r r
                r r
                r r
                r r}
\toprule
\multirow{2}{*}{\textbf{Model}}  & \multirow{2}{*}{\textbf{Seqlen}}
& \multicolumn{2}{c}{\textbf{\texttt{aten::*mm}}}
& \multicolumn{2}{c}{\textbf{\texttt{aten::softmax}}}
& \multicolumn{2}{c}{\textbf{\texttt{aten::*norm}}}
& \multicolumn{2}{c}{\textbf{\texttt{aten::*lu}}} \\
\cmidrule(lr){3-4}\cmidrule(lr){5-6}\cmidrule(lr){7-8}\cmidrule(lr){9-10}
 &  & \textbf{Time (ms)} & \textbf{Time (\%)} & \textbf{Time (ms)} & \textbf{Time (\%)} & \textbf{Time (ms)} & \textbf{Time (\%)} & \textbf{Time (ms)} & \textbf{Time (\%)} \\
\midrule
\multirow{4}{*}{Llama-3.2-3B}  & 128  & 6.336 & 9.40\% & 0.129 & 0.19\% & 4.154 & 6.16\% & 0.120 & 0.18\% \\
&  256  & 6.333 & 9.68\% & 0.129 & 0.20\% & 4.071 & 6.22\% & 0.119 & 0.18\% \\
&  512  & 6.358 & 9.69\% & 0.130 & 0.20\% & 3.959 & 6.03\% & 0.121 & 0.18\% \\
&  1024 & 6.356 & 9.47\% & 0.130 & 0.19\% & 4.072 & 6.07\% & 0.121 & 0.18\% \\
\midrule
\multirow{4}{*}{bert-base}  &  128  & 1.439 & 8.64\% & 0.079 & 0.48\% & 0.304 & 1.83\% & 0.047 & 0.28\% \\
&  256  & 1.438 & 8.60\% & 0.079 & 0.48\% & 0.304 & 1.82\% & 0.047 & 0.28\% \\
&  512  & 1.445 & 7.07\% & 0.079 & 0.39\% & 0.304 & 1.49\% & 0.047 & 0.23\% \\
&  1024 & 1.441 & 8.20\% & 0.079 & 0.45\% & 0.304 & 1.73\% & 0.047 & 0.27\% \\
\bottomrule
\end{tabular}
\label{tab:profile-decode}
}
\end{table*}

%% file: sec/3_baseline.tex
\section{Building a binary weight and ternary activation Transformer Baseline} 
\label{sec:baseline_scheme}

Since there is no quantization scheme for BWTA bitwidth before, we build a foundational framework for BWTA Transformers by incorporating existing binary and ternary functions and constructing basic Linear and MatMul layers. 

\subsection{Binary and Ternary Functions}

\subsubsection{Binary Functions}
We first introduce the binarization operators involved in our framework. We apply the $\operatorname{sign}$ function for weight $w$, where the forward propagation function is
\begin{equation}
\label{eq:sign}
    \operatorname{sign}(w) = 
            \begin{cases}
            1,  & \text{ if } w \ge 0\\
            -1, & \text{ otherwise }
            \end{cases}. 
\end{equation} 
We use the straight-through estimator (STE) to obtain the gradients in the backward propagation as a common practice~\cite{bengio2013estimating}. 
As for all non-negative activation (i.e., ReLU outputs and attention probability), we follow BiBERT~\cite{qin2022bibert} to binarize it to boolean values using 
\begin{equation}
    \operatorname{bool}(a, s_a) = 
    \begin{cases}
        1,  & \frac{a}{s_a} \ge 0.5 \\
        0,  &  \frac{a}{s_a} < 0.5
    \end{cases}.
\end{equation}
where $a$ is the activation, $s_a\in\mathbb{R}^{+}$ is the scaling factor for $a$. It can be reformulated to $\left \lfloor \operatorname{clip}(\frac{a}{s_a}, 0, 1) \right \rceil$ following~\cite{esser2019lsq} to enable learnable scaling factors. 

\subsubsection{Ternary Function} 
Except non-negative activations, we apply $\operatorname{ternary}$ function for the other activations that obey the Gaussian-distribution, which can be written as
\begin{equation}
    \operatorname{ternary}(a, s_a) = \begin{cases}
        1, &  \frac{a}{s_a} \ge 0.5 \\ 
        0, &  -0.5 \le \frac{a}{s_a} < 0.5 \\
        -1, &  \frac{a}{s_a} < -0.5 
    \end{cases}.
\end{equation}
Same as above, it is regarded as $\left \lfloor \operatorname{clip}(\frac{a}{s_a}, -1, 1) \right \rceil$ to enable stable training for the scalings. 

\subsection{Basic Layers for Linear and MatMul}

We redistribute the weight matrix $W$ to zero-mean to retain the information and apply scaling factors $s_W\in\mathbb{R}^+$ to align the magnitude between floating-point (FP) numbers and quantized ones to minimize the quantization error
\begin{equation}
\label{eq:bw}
W_{\operatorname{sign}} = s_W \cdot \operatorname{sign}(W - \mu({W})), \quad s_{W} = \frac{1}{n_{W}} ||{W}||_{F},
\end{equation}
where ${W}$ and $W_{\operatorname{sign}}$ denote full-precision and binarized weight respectively, $\mu(\cdot)$ denotes the mean value, $F$ means Frobenius norm. 

For activation, we do not apply mean-shift to reduce the runtime computation during training, and thus 
\begin{equation}
\label{eq:Aq}
    A_q = s_A \cdot \mathrm{quant}(A, s_A), \quad 
    s_{A}^0 = \frac{2}{n_{A}} ||{A}||_{\ell 1},
\end{equation}
where $\mathrm{quant}\in \{\operatorname{bool}$, $\operatorname{ternary}\}$ means non-negative activation and other activations, respectively. ${A}$ and $A_q$ are full-precision and quantized activations, $s_{A}^0$ is the initialization of scaling factor at the very beginning of the whole training. $\ell 1$ means absolute value. 

Therefore, the BWTA linear layer can be expressed as 
{\footnotesize
\begin{align}
\label{eq:bwta_linear}
    \operatorname{linear}(A) & = s_W s_A  \left(\operatorname{sign}({W} - \mu({W})) \otimes \operatorname{quant}({A}^\top, s_A)  \right) \\
    & = s_W s_A \cdot 
    \operatorname{BWTA}\left({W} - \mu({W}), \frac{A^\top}{s_A}\right),
\end{align}}
where $\otimes$ represents the Matrix Multiplication and Accumulation (MMA) operator implemented with bitwise instructions, such as logical gates \texttt{xor}, \texttt{and}, and accumulation \texttt{popcount} on hardware, but is usually kept as full-precision GEMM during training. We design $\operatorname{BWTA}(\cdot, \cdot)$ to facilitate the general linear layers on GPUs, loading binary weight in memory and transforming activation to ternary for multiplication.  

In the Attention mechanism of Transformer models, there are two MatMuls that receive different inputs. The first rule is  used to compute the attention scores by $Q$ and $K$: 
{\footnotesize
\begin{align}
\label{eq:bwta_qk}
    Att_{score} & = \frac{s_Q s_K}{\sqrt{D}} \left( \operatorname{ternary}(Q, s_Q) \otimes \operatorname{ternary} (K^\top, s_K) \right) \\
    & = \frac{s_Q s_K}{\sqrt{D}} \cdot 
 \operatorname{BWTA}_{QK} \left( \frac{Q}{s_Q}, \frac{K^\top}{s_K} \right),
\end{align}}
where $s_Q$ and $s_K$ are learnable layerwise scaling factors for query and key matrices, $D$ is the hidden state size. $\operatorname{BWTA}_{QK}(\cdot, \cdot)$ is the custom MatMul kernel to transform both inputs to ternary representation and then conduct multiplication. In Transformer, there is only query and key obey this arithmetic rule, so we use $_{QK}$ as subscript to make it distinct from $\operatorname{BWTA}(\cdot, \cdot)$. 

And the other one is the multiplication of attention probability $Att_{prob}$ (shorten as $Att$ in the following text) and value $V$. 
{\footnotesize
\begin{align}
    Context  & = s_{Att}s_V \left( \operatorname{bool} (Att, s_{Att}) \otimes \operatorname{ternary} (V, s_V) \right) \\
     & = s_{Att}s_V \cdot \operatorname{BWTA}_{Att} \left( \frac{Att}{s_{Att}}, \frac{V}{s_V}\right),
\end{align}}
where $s_{Att}$ and $s_V$ are learnable scaling factors for $Att$ and $V$, respectively. $\operatorname{BWTA}_{Att}(\cdot, \cdot)$ is the custom MatMul kernel to transform the $Att$ and $V$ to boolean and ternary respectively and then compute the results. 

We follow original Transformer architecture to construct the Multi-Head Attention (MHA) and Feed-Forward Network (FFN) with BWTA layers and MatMuls, leaving the first and last layers full precision to keep the accuracy.

%% file: sec/appendix/exp_details.tex
\subsection{Configuration Details}
\textbf{Training Configurations on BERT. } 
We apply the clipped STE for activations and  the naive STE for weights in the backpropagation. 
And we adopt a common distillation scheme, leveraging intermediate states, FFN output, and prediction logits as knowledge~\cite{liu2022bit, hou2020dynabert}. Additionally, we utilize the distillation loss function in \cite{martinez2020trainingr2b} on attention probabilities in each block. 
As for comparison methods, we follow their original training procedures and settings, some of which use multi-stage training schemes and specific distillations. 

Since we leverage the multi-stage quantization scheme for our method, the overall training schedule is generated at the beginning. Usually, we set half of the epochs for ternary stage, and evenly divide the rest epochs for the earlier stages. We take different training epochs according to the volume of the dataset to ensure sufficient training. The total epochs for each task are $80$ for CoLA, $30$ for MRPC, STS-B and RTE, $20$ for SST-2 and QNLI, and $6$ for MNLI and QQP. 

We adopt AdamW optimizer with 0.01 weight decay. The initial learning rate is 5e-4, and we use a cosine annealing schedule with 5 epochs of warm-up. We set the learning rate to $1e-3$ and $2e-5$ for the learnable scaling factors and the weight respectively, with $16$ as batch size. At the stage transition, the optimizer will be initialized. 

\new{\textbf{Training Configurations on LLM. } In practice, jointly pushing both weights and activations to ultra-low bit-widths can noticeably hurt LLM performance; however, we find that selectively quantizing less-sensitive layers is feasible and yields favorable trade-offs. Our additional experiments in our revision is designed around this insight. }

\new{We first conduct layer selection on the target LLM. We use a small calibration set to identify least-sensitive linear layers by their post-quantization MSE and select those with the smallest average error. We replace the selected layers with our BWTA modules (1-bit weight and ternary activation) based on the pretrained Bitnet model.}

\new{We train these BWTA layers with our smooth multi-stage framework (SMQ) to ensure stable convergence under ultra-low bits. Concretely, we set the stages to $L=\{19,15,11,7,3\}$ with a linearly decreasing stride to gradually tighten activation bins while preserving training stability. Each stage uses only 1-2k steps with AdamW optimzier, use early-stop strategy for the beginning stages based on the evaluation loss. (Note that Bitnet uses 40k steps to train models of each size. Since we start from a low-bit quanized pretrained weights instead of an FP16 model, we shrink the training steps.) We warmup the training in the early 3\% steps, and the initial learning rate is 2e-4, weight decay is set to 0.} 

\new{In final, we perform end-to-end timing in both prefill and decode phases under diverse workload settings (various batch sizes, context lengths, and generation lengths). Meanwhile, we evaluate perplexity on WikiText-2 and C4, and the model accuracy on CommonsenseQA benchmarks to measure the impact of ultra low-bit quantization on generation quality. }


%% file: sec/appendix/gradient_of_scaling.tex
\subsection{Further Analysis on the Gradients of Scaling Factors}

\begin{figure}[t]
    \centering
    \includegraphics[width=0.97\linewidth]{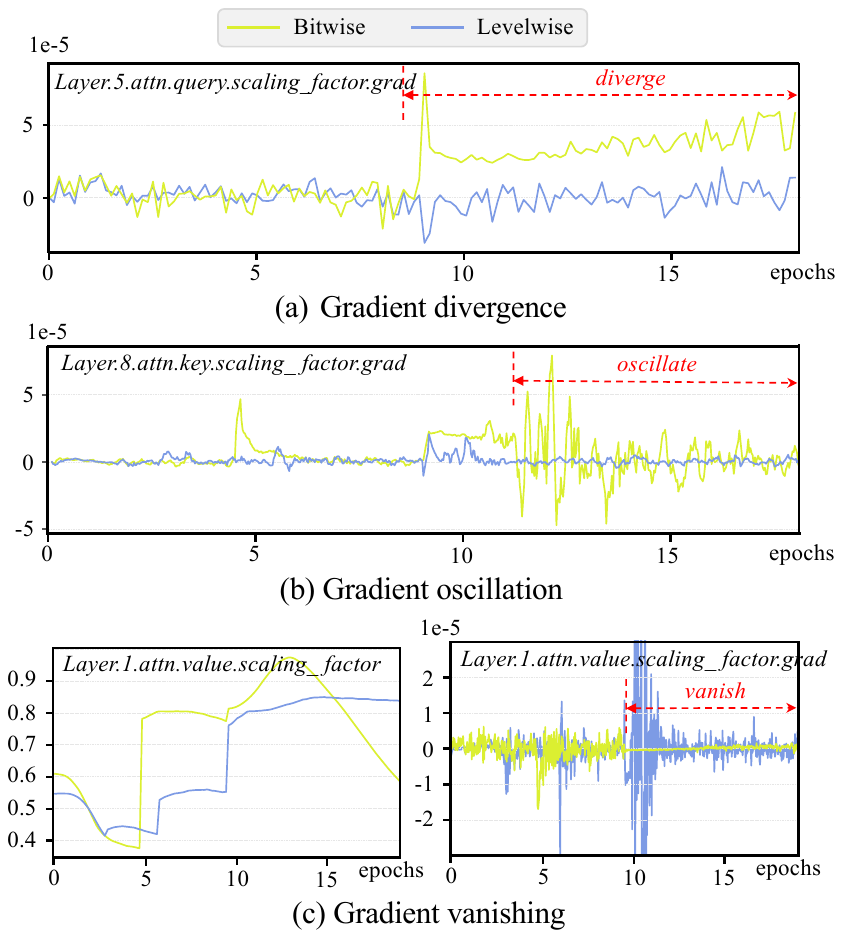}
    \caption{The curves of scaling factors and their gradients. }
    \label{fig:visualization_gradient}
\end{figure}

By tracing the scaling factors and the gradients during training, we find that our Smooth Multi-stage Quantization framework improves the convergence of the scaling factors within limited epochs. 

Fig.~\ref{fig:visualization_gradient} illustrates three issues in gradients that may arise during training: 
(a) \textit{Gradient divergence}. The gradient of the bitwise degradation strategy is easier to diverge on one side of zero along with the training epochs, especially after transiting to the binary stage. See Fig.~\ref{fig:visualization_gradient}(a), as the gradient remains positive, the scaling factor continuously decreases and fails to converge to an optimum value. 
(b) \textit{Gradient oscillation}.  In this case, the gradient keeps oscillating till the end of the training, making the scaling factor oscillate or hard to converge (Fig.~\ref{fig:visualization_gradient}(b)). 
(c) \textit{Gradient vanishing}. It often occurs in shallow layers, and arises when the gradient becomes rather small before the scaling factor converges to a local optimum. Since the AdamW optimizer calculates the weighted average of the past gradients through momentum, it tends to maintain the previous movement to update the parameter. As a result, even when the gradient vanishes, the scaling factor continuously updates in the previous direction without reaching an optimal value.

Therefore, due to the aforementioned gradient issues, stable training and fast convergence of ultra-low bit BERT model are particularly difficult.  
In TABLE~\ref{tab:not_converge_proportion}, we quantify the proportion of the scaling factors exhibiting non-convergent behaviors across various tasks. In bitwise degradation strategy, the proportion of the scaling factors fail to converge is even more than 30\% in QNLI, STS-B, CoLA, and MRPC. However, levelwise degradation strategy alleviates the issue, demonstrating that the proposed framework effectively mitigates the occurrence of the gradient issues, enabling fast convergence of ultra-low bit models within the limited training time. 

\input{tables/scaling_factors_not_converged}

%% file: tables/scaling_factors_not_converged.tex
\begin{table}[h]
    \centering
    \resizebox{0.95\textwidth}{!}{
    \begin{tabular}{cccccc}
    \toprule
      \textbf{Strategy} &  \textbf{QNLI} &  \textbf{SST-2} & \textbf{STS-B} & \textbf{CoLA} & \textbf{MRPC}\\ \hline
      Bitwise (\%)   & 34.2 & 28.1 & 30.2 & 31.5 & 37.0 \\  \hline
      Levelwise (\%) & 11.6 & 14.4 & 13.7 & 10.9 & 19.8 \\ 
    \bottomrule
    \end{tabular}
    }
    \caption{Scaling factors not converged within the pre-defined training epochs. }
    \label{tab:not_converge_proportion}
\end{table}

%% file: sec/appendix/exp_llm.tex
\subsection{More Experimental Results on GLUE Benchmark}

\input{tables/accuracy_glue_wo_DA}
\label{app:glue-wo-DA}
We provide the results without data augmentation in TABLE~\ref{tab:glue-wo-DA}. All the comparison methods are followed their official released codes and papers. We can see that our BWTA achieves the highest accuracy on all the tasks among all the binarization methods.

\input{tables/accuracy_cmQA}
\label{app:exp_llm}
\new{\subsection{More Experimental Results on CommonsenseQA Benchmark}
We show accuracy on commonsenseQA benchmarks in Table~\ref{tab:accuracy-llm-cmQA}. We compare our method with low-bit quantization methods, including RTN (round-to-nearest), AWQ~\cite{lin2024awq}, GPTQ~\cite{gptq}, OmniQuant~\cite{shao2023omniquant}, and  and the recent 1-bit LLM works, including DB-LLM~\cite{chen2024dbllm}, PB-LLM~\cite{shang2023pbllm}, BiLLM~\cite{huang2024billm} and Bitnet~\cite{wang2023bitnet}. 
Results show that our BWTA achieves comparable accuracy even with lower average bitwidth and model sizes, indicating negligible degradation in generation quality. }

%% file: tables/accuracy_glue_wo_DA.tex
\begin{table*}[ht]
\caption{Comparison of BERT quantization methods without data augmentation. ``--'' means the results are not officially reported. }
\label{tab:glue-wo-DA}
{\footnotesize
\centering
\resizebox{\textwidth}{!}{
\begin{tabular}{llccccccccccc}
\toprule
\textbf{Quant}  & \begin{tabular}[c]{@{}c@{}}\textbf{\#Bits}\end{tabular} & \begin{tabular}[c]{@{}c@{}}\textbf{Size}$_\text{ (MB)}$\end{tabular}  & \begin{tabular}[c]{@{}c@{}}\textbf{MNLI}$_\text{-m/mm}$\end{tabular} & \textbf{QQP} & \textbf{QNLI} & \textbf{SST-2} & \textbf{CoLA} & \textbf{STS-B} & \textbf{MRPC} & \textbf{RTE} & \textbf{Avg.} \\ 
\midrule
Full Precision & 32/32 & 418 & 84.9/85.5 & 91.4 & 88.6 & 93.2 & 59.7 & 89.8 & 86.2 & 72.2  & 83.9 \\ 
\midrule
Q2BERT  &  2/8 & 43.0 & 47.2/47.3 & 67.0 & 61.3 & 80.6 & 0 & 4.4 & 68.4 & 52.7 &  47.7 \\
Q-BERT & 2/8 & 43.0 & {76.6/77.0} & -- & -- & {84.6} & -- & -- & {68.3} & {52.7} & -- \\
TernaryBERT & 2/8 & 28.0 & 83.3/83.3 & 90.1 & -- & -- & 50.7 & -- & 87.5 & 68.2 & -- \\
TernaryBERT & 2/2 & 28.0 & 40.3/40.0 & 63.1 & 50.0 & 80.7 & 0 & 12.4 & 68.3 & 54.5 &  45.5 \\
{TernaryBERT} & 2/1 & 28.0 & 32.7/33.0 & 74.1 & 59.3 & 53.1 & 0 & 7.1 & 68.3 & 53.4 & 42.3 \\
BinaryBERT &  1/4 & 16.5 & 83.9/84.2 & 91.2 & 90.9 & 92.3 & 44.4 & 87.2 & 83.3 & 65.3 & 79.9 \\
\new{BEBERT} & 1/4 & 33.0 & 84.7/84.7 &  91.4 &  91.1 &  93.2 & 54.0 & -- &  84.6  &  68.8  & 80.5 \\ 
BinaryBERT & 1/2 & 16.5  & 62.7/63.9 & 79.9 & 52.6 & 82.5 & 14.6 & 6.5 & 68.3 & 52.7 &  53.7 \\
\new{BiPFT} &  1/2 &  14.9 &  77.0/76.9 &  89.0 &  86.6 &  88.1 & 36.2 &  87.0 &  84.1 &  66.1 &  76.8\\
\midrule
BinaryBERT & 1/1 & 16.5 & 35.6/35.3 & 66.2 & 51.5 & 53.2 & 0 & 6.1 & 68.3 & 52.7 & 41.0 \\
BiBERT & 1/1 & 13.4 & 66.1/67.5 & 84.8 & 72.6 & 88.7 & 25.4 & 33.6 & 72.5 & 57.4 & 63.2\\
BiT & 1/1 & 13.4  & 79.5/79.4 & 85.4 & 82.4 & 89.9 & 32.9 & 72.0 & 79.9 & 62.1 & 73.5 \\
\new{BEBERT} & 1/1 & -- &  75.7/76.6 &  84.9 &  80.7 &  90.2 & 27.7 & -- &  75.1  &   58.6  & 71.2 \\ 
\new{BiPFT} & 1/1 & 14.9 &  69.5/70.6  &  83.7 &  81.7  &  86.2 &  22.9  &  80.2  &  76.2 &  66.1 &  70.8 \\
\textbf{Ours} &  \textbf{1/1.5} &  \textbf{13.4} & \textbf{81.0/81.3} & \textbf{89.2} & \textbf{82.9} & \textbf{91.5} & \textbf{33.5} & \textbf{79.0} & \textbf{81.3} & \textbf{64.3} & \textbf{76.2} \\
\bottomrule
\end{tabular}
}}
\end{table*}

%% file: tables/accuracy_cmQA.tex
\begin{table*}[ht]
    \centering
    \caption{Accuracy on commonsense QA benchmarks for various models and quantization methods.}
    \label{tab:accuracy-llm-cmQA}
    \resizebox{\textwidth}{!}{
    \begin{tabular}{l l c r r r r r r r r}
    \toprule
    \textbf{Model} & \textbf{Method} & \textbf{W$_{bit}$/A$_{bit}$} &
    \textbf{BoolQ} & \textbf{PIQA} & \textbf{HS} &
    \textbf{WG} & \textbf{ARC-E} & \textbf{ARC-C} &
    \textbf{OBQA} & \textbf{Avg.} \\
    \midrule
    \multirow{3}{*}{LLaMA-2-7B}
      & Full Precision            & 16/16 & 77.4 & 78.8 & 77.2 & 69.2 & 75.2 & 45.9 & 58.6  & 68.9 \\
      & GPTQ          & 2/16 & 43.9 & 51.1 & 26.3 & 50.8 & 26.6 & 28.5 & 29.0 & 36.6 \\
      & PB-LLM        & 1.7/16  & 62.3 & 53.8 & 27.7 & 49.3 & 28.0 & 25.0 & 30.2 & 39.5 \\
      & BiLLM         & 1.1/16  & 61.8 & 60.6 & 34.8 & 52.4 & 36.2 & 24.4 & 33.2 & 43.3 \\
    \midrule
    \multirow{5}{*}{Bitnet-b1.58-3B}
      & Bitnet        & 1.5/8   & 58.9 & 72.3 & 42.8 & 60.8 & 65.0 & 30.0 & 26.6 & 50.9 \\
      & BWN \textit{(10\% layers)}         & 1.4/7 & 37.8 & 49.6 & 25.0 & 49.6 & 25.7 & 22.9 & 27.0 & 33.9   \\
      & \textbf{BWTA (Ours) \textit{(10\% layers)}} & \textbf{1.4/7} & \textbf{63.7} & \textbf{71.9} & \textbf{42.0} & \textbf{59.9} & \textbf{65.5} & \textbf{29.4} & \textbf{26.6} & \textbf{51.3}  \\
     & BWN \textit{(30\% layers)}          & 1.3/6    \\
     & \textbf{BWTA (Ours) \textit{(30\% layers)}} & \textbf{1.3/6} & \\  
    \midrule
    
    \multirow{5}{*}{Bitnet-b1.58-1.3B}
      & Bitnet   & 1.5/8  & 62.0 & 68.9 & 37.5 & 55.6 & 59.3 & 26.2 & 20.6 & 47.2 \\
      & BWN \textit{(10\% layers)}  & 1.4/7 & 59.9 & 69.3 & 37.1 & 55.0 & 57.8 & 26.3 & 20.8 & 46.6  \\
    & \textbf{BWTA (Ours) \textit{(10\% layers)}} & \textbf{1.4/7} &  \textbf{61.5} & \textbf{68.4} & \textbf{37.6} & \textbf{55.5} & \textbf{59.0} & \textbf{26.2} & \textbf{22.4} & \textbf{47.2}  \\ 
    
    & BWN \textit{(30\% layers)}  & 1.3/6 & 59.8 & 62.8 & 32.8 & 51.5 & 48.0 & 22.4 & 20.4 & 42.5   \\
    & \textbf{BWTA (Ours) \textit{(30\% layers)}} & \textbf{1.3/6} & \textbf{57.5} & \textbf{66.5} & \textbf{35.4} & \textbf{53.7} & \textbf{54.0} & \textbf{25.2} & \textbf{21.0} & \textbf{44.8} \\ 
    
    \midrule
    
    \multirow{5}{*}{Bitnet-b1.58-0.7B} 
      & Bitnet     & 1.5/8   & 54.0 & 67.9 & 35.0 & 53.9 & 55.4 & 24.7 & 20.0 & 44.4 \\
      & BWN \textit{(10\% layers)}     & 1.4/7  & 52.8 & 65.9 & 32.9 & 51.2 & 52.5 & 23.6 & 19.0 & 42.6  \\
      & \textbf{BWTA (Ours) \textit{(10\% layers)}} & \textbf{1.4/7} &  \textbf{52.6} & \textbf{68.3} & \textbf{34.7} & \textbf{54.0} & \textbf{54.2} & \textbf{25.4} & \textbf{22.0} & \textbf{44.5}  \\
     & BWN \textit{(30\% layers)}  & 1.3/6 & 53.7 & 62.5 & 32.1 & 49.8 & 43.9 & 21.5 & 19.0 & 40.4   \\
    & \textbf{BWTA (Ours) \textit{(30\% layers)}} & \textbf{1.3/6} & \textbf{58.2} & \textbf{65.0} & \textbf{33.0} & \textbf{52.1} & \textbf{50.2} & \textbf{23.5} & \textbf{22.0} & \textbf{43.4} \\ 
    \bottomrule
\end{tabular}}
\end{table*}

%% file: sec/appendix/more_results_of_kernel_benchmark.tex
\subsubsection{Ablation Studies on Kernel Design}
\label{app:ablation-kernel-speed}
\input{tables/exp_bitpacking}
\input{tables/exp_mma}

We conduct speed evaluations for our binary/ternary bitpack (TABLE~\ref{tab:speed_bitpacking}) and the BWTA MatMul kernel (TABLE~\ref{tab:speed_matmul_kernel}) separately. 
We selected typical shapes frequently used in Transformer-based architectures, specifically for Linear and MatMul operations in BERT-base and Llama~\cite{touvron2023llama}. The results include the latency  ($\mu$s) and TFLOPs for each case, allowing for a comprehensive evaluation of performance across varying input dimensions.

In TABLE~\ref{tab:speed_bitpacking}, we evaluate the performance of the BWTA layer with once Ternary Bitpack for activation in linear layers, and BWTA$_{QK}$ with twice Ternary Bitpack for both query and key matrices. The MatMul operation is omitted to avoid the effects of the computational steps. 
{Naive} means one-by-one bitpacking using \texttt{setp} instruction, which requires two \texttt{setp} and two conditional executions for each value, resulting in a double number of instructions compared to BWTA. The results in the table indicate that our BWTA consistently outperforms the naive implementation across all the shapes of input matrices. The speedup is particularly pronounced with larger matrices. For example, BWTA is 0.0056 $\mu s$ faster than {Naive} with $13824\times 5120$ and $5120\times32$ matrices, which is a typical matrix shape in LLMs. 

In TABLE~\ref{tab:speed_matmul_kernel}, we also compare our BWTA to {Naive} implementation of MatMul, which utilizes four \texttt{mma.and} operations. Since our BWTA kernel employs only two \texttt{mma.xor} and one \texttt{sub}, resulting in less than half the number of instructions required by the Naive implementation (see Fig.~\ref{fig:low_level_perspective} \textit{Case 1} for details). Consequently, the latency of our BWTA kernel is obviously lower. Compared to the latency of {w/o}, which refers to the kernel without computational steps, we find that the time taken by the MatMul kernel in BWTA is approximately half of that taken by the MatMul kernel in the {Naive} implementation.

\subsection{Results of Kernel Benchmark with More Shapes}
\label{app:more_results_of_kernel_benchmark}

\input{tables/exp_kernel_latency_shapes}

We test the latency for three variants of BWTA MatMul kernels with diverse shapes. \new{Results shown in TABLE~\ref{tab:latency_comparison} are tested on NVIDIA  A800 with 80GB VRAM.} Since the different layers in Transformer-based models have various shapes, we take $\{128, 512, 768, 2048, 3072\}$ in our experiments, as they are commonly used. We compare the Linear, MatMul, and Attention layers separately to show the efficiency of different variants of BWTA in each case. The rows of FP16 mean the kernel with the same data pipeline and thread allocation as our BWTA, but uses the 16-bit floating-point \texttt{mma} instruction to compute the MatMul results. The rows of cuBLAS mean we directly call the GEMM API $\operatorname{cublasHgemm}$ provided by cuBLAS~\footnote{https://docs.nvidia.com/cuda/cublas} to compute in half-precision. 

We can see from TABLE~\ref{tab:latency_comparison} that the speed of our BWTA far exceeds the floating-point counterparts. In $W\otimes A^\top$ case, our BWTA kernel shows a leading performance, being approximately 23-35$\times$ faster than the FP16 implementation, and achieving a 4.5$\times$ speedup compared to cuBLAS HGEMM. It indicates that the acceleration provided by our BWTA is primarily attributed to the ultra-low bit computational operations and compact parameters, presenting the potential for broader applications that require low latency. The BWTA kernels in $Att \otimes V$ and $Q\otimes K^\top$ cases exhibit a similar acceleration ratio. While further improvements can be made by optimizing the data allocation and thread parallelism to extend the speedup ratio and resource utilization, the acceleration potential has been proved.

%% file: tables/exp_bitpacking.tex

\begin{table}[]
    \centering
    {\footnotesize
    \begin{tabular}{llcc}
    \toprule
    Shape & Method & Latency ($\mu$s) & TFLOPs \\   \hline
    \multicolumn{4}{l}{BWTA \textit{with once Ternary Bitpack}} \\
       \multirow{3}{*}{ 
        \begin{tabular}{l}
            $W$: $[128, 128]$  \\
            $A$: $[128, 128]$  
        \end{tabular}}       & w/o    & 8.31  &  0.505 \\ 
                             & Naive  & 10.47  &  0.398 \\ 
                             & IPB   & \textbf{10.01}  &  \textbf{0.416} \\ \hline

         \multirow{3}{*}{ 
        \begin{tabular}{l}
            $W$: $[768, 768]$  \\
            $A$: $[128, 768]$  
        \end{tabular}}       & w/o    & 7.59  &  19.90 \\ 
                             & Naive  & 11.38  &  13.26 \\ 
                             & IPB   & \textbf{10.93}  &  \textbf{13.78} \\ \hline

        \multirow{3}{*}{ 
        \begin{tabular}{l}
            $W$: $[3072, 768]$  \\
            $A$: $[128, 768]$  
        \end{tabular}}       & w/o    & 9.64  &  62.37 \\ 
                             & Naive  & 14.22  &  42.53 \\ 
                             & IPB   & \textbf{13.83}  &  \textbf{43.71} \\ \hline

         \multirow{3}{*}{ 
        \begin{tabular}{l}
            $W$: $[13824, 5120]$  \\
            $A$: $[32, 5120]$  
        \end{tabular}}       & w/o    & 35.02  & 129.12 \\ 
                             & Naive  & 67.24  & 67.32 \\ 
                             & IPB   & \textbf{61.56}  & \textbf{73.45} \\ \hline

        \multicolumn{4}{l}{BWTA$_{QK}$ \textit{with twice Ternary Bitpack}} \\
         
         \multirow{3}{*}{ 
        \begin{tabular}{l}
            $Q$: $[1024, 1024]$  \\
            $K$: $[1024, 1024]$  
        \end{tabular}}       & w/o    & 18.53  &  115.97 \\ 
                             & Naive  & 19.14  &  112.52 \\ 
                             & IPB   & \textbf{18.89}  &  \textbf{113.30} \\ \hline

         \multirow{3}{*}{ 
        \begin{tabular}{l}
            $Q$: $[2048, 2048]$  \\
            $K$: $[2048, 2048]$  
        \end{tabular}}       & w/o    & 115.7  &  148.48 \\ 
                             & Naive  & 117.2  &  146.55 \\ 
                             & IPB   & \textbf{116.3}  &  \textbf{147.65} \\ \hline

    \bottomrule
    \end{tabular}}
    \caption{Speed ablation of bitpacking implements. w/o means skip the bitpack operation, Naive means one-by-one bitpacking with \texttt{setp} instruction. IPB is our Instruction-level Parallel Bitpack method using \texttt{set} instruction. } 
    \label{tab:speed_bitpacking}
\end{table}

%% file: tables/exp_mma.tex

\begin{table}[]
    \centering
    {\footnotesize
    \begin{tabular}{llcc}
    \toprule
    Shape & Method & Latency ($\mu$s) & TFLOPs \\   \hline
      \multirow{3}{*}{ 
        \begin{tabular}{l}
            $W$: $[128, 128]$  \\
            $A$: $[128, 128]$ 
        \end{tabular}}        
                              & w/o                      & 5.44 & 0.772 \\ 
                              & Naive (\texttt{mma.and}) & 7.46 & 0.562 \\  
                              & BWTA (\texttt{mma.xor})  & \textbf{6.96} & \textbf{0.603} \\ \hline
        
     \multirow{3}{*}{ 
        \begin{tabular}{l}
            $W$: $[768, 768]$  \\
            $A$: $[128, 768]$ 
        \end{tabular}}        
                              &  w/o                     & 10.91 & 13.84 \\
                              & Naive (\texttt{mma.and}) & 13.75 & 10.98 \\  
                              & BWTA (\texttt{mma.xor})  & \textbf{11.33} & \textbf{13.33} \\ \hline
    \multirow{3}{*}{ 
        \begin{tabular}{l}
            $W$: $[3072, 768]$  \\
            $A$: $[128, 768]$ 
        \end{tabular}}        
                              & w/o                      & 12.52 & 48.25 \\
                              & Naive (\texttt{mma.and}) & 13.24 & 45.63 \\  
                              & BWTA (\texttt{mma.xor})  & \textbf{12.88} & \textbf{46.89}  \\ \hline
                              
    \multirow{3}{*}{ 
        \begin{tabular}{l}
            $W$: $[13824, 5120]$  \\
            $A$: $[32, 5120]$ 
        \end{tabular}}        
                              & w/o                      & 53.78 & 84.23 \\
                              & Naive (\texttt{mma.and}) & 58.58 & 77.32 \\  
                              & BWTA (\texttt{mma.xor})  & \textbf{55.08} & \textbf{82.24}  \\ \hline
    \bottomrule
    \end{tabular}}
    \caption{Speed ablation of MatMul kernels. \textit{w/o} means we skip the MatMul operation. \textit{Naive} is a straightforward implementation without optimizing. BWTA is our efficient \texttt{xor}-based BWTA MatMul.  }
    \label{tab:speed_matmul_kernel}
\end{table}

%% file: tables/exp_kernel_latency_shapes.tex
\begin{table}[]
    \centering
    {\footnotesize
    \begin{tabular}{lllc}
    \toprule
    Layer & Shape & Kernel & Latency ($\mu$s) \\  \hline 
    \multirow{9}{*}{\begin{tabular}{c}
      $W \otimes A^\top$ \\ (Linear)
    \end{tabular}
    }   & \multirow{3}{*}{ 
                                        \begin{tabular}{l}
                                            $W$: $[768, 768]$  \\
                                            $A$: $[128, 768]$ 
                                        \end{tabular}}                  & FP16   &  215.1  \\ 
                                     &                                  & cuBLAS &  41.32  \\ 
                                     &                                  & BWTA   &  \textbf{9.24} \\ \cline{2-4}
                                     
                                     & \multirow{3}{*}{ 
                                        \begin{tabular}{l}
                                            $W$: $[3072, 768]$  \\
                                            $A$: $[128, 768]$ 
                                        \end{tabular}}                  & FP16   & 216.3   \\ 
                                     &                                  & cuBLAS & 80.17   \\ 
                                     &                                  & BWTA   & \textbf{10.43}  \\ \cline{2-4}

                                     & \multirow{3}{*}{ 
                                        \begin{tabular}{l}
                                            $W$: $[768, 3072]$  \\
                                            $A$: $[128, 3072]$ 
                                        \end{tabular}}                  & FP16   &  850.8  \\ 
                                     &                                  & cuBLAS &  105.6 \\ 
                                     &                                  & BWTA   &  \textbf{24.38} \\ \hline

    \multirow{9}{*}{$Att \otimes V$} & \multirow{3}{*}{ 
                                        \begin{tabular}{l}
                                            $Att$: $[128, 128]$  \\
                                            $V$: $[128, 128]$ 
                                        \end{tabular}}                  & FP16   & 51.12  \\
                                     &                                  & cuBLAS & 38.68 \\ 
                                     &                                  & BWTA$_{Att}$ & \textbf{6.38} \\  \cline{2-4}

                                     & \multirow{3}{*}{ 
                                        \begin{tabular}{l}
                                            $Att$: $[512, 512]$  \\
                                            $V$: $[512, 512]$ 
                                        \end{tabular}}                  & FP16 & 145.9   \\
                                     &                                  & cuBLAS & 57.42 \\ 
                                     &                                  & BWTA$_{Att}$ & \textbf{9.77} \\  \cline{2-4}

                                     & \multirow{3}{*}{ 
                                        \begin{tabular}{l}
                                            $Att$: $[2048, 2048]$  \\
                                            $V$: $[2048, 2048]$ 
                                        \end{tabular}}                  & FP16   & 2833 \\
                                     &                                  & cuBLAS & 611.9 \\ 
                                     &                                  & BWTA$_{Att}$ & \textbf{201.6} \\  \hline
    
    \multirow{9}{*}{$Q \otimes K^{\top}$}  & \multirow{3}{*}{ 
                                        \begin{tabular}{l}
                                            $Q$: $[128, 128]$  \\
                                            $K$: $[128, 128]$ 
                                        \end{tabular}}                  & FP16          &  39.31  \\ 
                                     &                                  & cuBLAS        &  52.56   \\ 
                                     &                                  & BWTA$_{QK}$   &  \textbf{5.41} \\ \cline{2-4}
                                     
                                     & \multirow{3}{*}{ 
                                        \begin{tabular}{l}
                                            $Q$: $[512, 512]$  \\
                                            $K$: $[512, 512]$ 
                                        \end{tabular}}                  & FP16          &  145.7  \\ 
                                     &                                  & cuBLAS        &  58.23   \\ 
                                     &                                  & BWTA$_{QK}$   &  \textbf{8.96} \\ \cline{2-4}

                                     & \multirow{3}{*}{ 
                                        \begin{tabular}{l}
                                            $Q$: $[2048, 2048]$  \\
                                            $K$: $[2048, 2048]$ 
                                        \end{tabular}}                  & FP16          &  2834  \\ 
                                     &                                  & cuBLAS        &  606.5   \\ 
                                     &                                  & BWTA$_{QK}$   &  \textbf{144.1} \\ \hline

    \bottomrule
    \end{tabular}
    }
    \caption{The latency comparison to the customized FP16 GEMM which has the same blocks and threads as our BWTA, and cuBLAS which has the optimized parallel management.}
    \label{tab:latency_comparison}
\end{table}

%% file: sec/appendix/limit.tex
\new{\section{Future Work}
\subsection{GPU-Only Kernel Design}
Our kernel is intentionally targeted at GPU/CPU to address today's dominant deployment stack for LLM training and inference. We chose GPUs for three reasons:
\begin{enumerate}
    \item Relevance \& comparability. GPUs are the de-facto platform for LLMs; evaluating on a single commodity platform enables fair, reproducible comparisons with existing low-bit baselines and real workloads (prefill/decode), without confounds from heterogeneous boards or toolchains.
    \item Co-design within real constraints. Our work co-optimizes BWTA with GPU realities (MMA tile geometry, register/shared-memory limits, low-bit instruction throughput, packing/layout). This is not software-only; the algorithmic choices were made because they map efficiently to GPU bitwise execution paths.  
    \item Community choice. GPU kernels can be immediately adopted by the most frameworks and toolkits, benefiting a wide range of models and inference pipelines. 
\end{enumerate}
We agree that FPGA/ASIC can further improve energy efficiency. However, our GPU-oriented kernels and layouts are not a 1:1 drop-in for FPGA/ASIC due to different instruction sets, memory hierarchies, and tool flows. Translating BWTA to custom silicon requires additional micro-architectural design (e.g., bit-serial/bit-parallel data paths, dsp-packing strategies, on-chip SRAM tiling, scale fusion), which is beyond the present scope. We leave it to our future works. }